\documentclass[preprint,12pt,authoryear,nonatbib]{elsarticle}

\usepackage{amssymb}
\usepackage{amsmath}
\usepackage{amsthm}

\journal{Expert Systems with Applications}

\usepackage[nolist,nohyperlinks]{acronym}
\acrodef{MLP}[MLP]{Multi-Layer Perceptron}
\acrodef{IoU}[IoU]{Intersection over Union}

\usepackage{booktabs}
\usepackage{multirow}
\usepackage{array}
\usepackage{makecell}

\usepackage{subcaption}

\usepackage[export]{adjustbox}

\usepackage{hyperref}

\usepackage{apacite}

\usepackage{mleftright}

\begin{document}

\begin{frontmatter}



  \title{Canonical Space Representation for 4D Panoptic Segmentation of Articulated Objects\tnoteref{funding}} 
  \tnotetext[funding]{
    This work was supported by FCT - Fundação para a Ciência e Tecnologia, I.P. under unit 00127-IEETA by 2023.01882.BD and \href{https://doi.org/10.54499/2023.01882.BD}{https://doi.org/10.54499/2023.01882.BD}
    and by "ERDF A Way of Making Europa", the Departament de Recerca i Universitats from Generalitat de Catalunya with reference2021SGR01499, and the Generalitat de Catalunya CERCA Program.
  }


  \author[IEETA]{Manuel Gomes\corref{cor1}} 
  \ead{manuelgomes@ua.pt}
  \author[CVC]{Bogdan Raducanu} 
  \ead{bogdan@cvc.uab.es}
  \author[IEETA]{Miguel Oliveira} 
  \ead{mriem@ua.pt}

  \cortext[cor1]{Corresponding author}

  \affiliation[IEETA]{organization={IEETA - LASI, Department of Mechanical Engineering, University of Aveiro},
    addressline={Campus Universitário de Santiago},
    city={Aveiro},
    postcode={3810-193},
  country={Portugal}}

  \affiliation[CVC]{organization={Computer Vision Center, Universitat Autònoma de Barcelona},
    addressline={Edifici O, Campus UAB},
    city={Barcelona},
    postcode={08193},
  country={Spain}}

  \begin{abstract}
    Articulated object perception presents significant challenges in computer vision, particularly because most existing methods ignore temporal dynamics despite the inherently dynamic nature of such objects.
    The use of 4D temporal data has not been thoroughly explored in articulated object perception and remains unexamined for panoptic segmentation.
    The lack of a benchmark dataset further hurt this field.
    To this end, we introduce Artic4D as a new dataset derived from PartNet Mobility and augmented with synthetic sensor data, featuring 4D panoptic annotations and articulation parameters.
    Building on this dataset, we propose CanonSeg4D, a novel 4D panoptic segmentation framework.
    This approach explicitly estimates per-frame offsets mapping observed object parts to a learned canonical space, thereby enhancing part-level segmentation.
    The framework employs this canonical representation to achieve consistent alignment of object parts across sequential frames.
    Comprehensive experiments on Artic4D demonstrate that the proposed CanonSeg4D outperforms state of the art approaches in panoptic segmentation accuracy in more complex scenarios.
    These findings highlight the effectiveness of temporal modeling and canonical alignment in dynamic object understanding, and pave the way for future advances in 4D articulated object perception.
  \end{abstract}


  \begin{highlights}
  \item Artic4D, a novel dataset which comprises 4D sensor data of articulated objects
  \item CanonSeg4D, the first panoptic 4D segmentation approach for articulated objects
  \item Usage of canonical representations to achieve articulation invariant segmentation
  \item Extensive experimental evaluation including comparison with state-of-the-art
  \end{highlights}


\end{frontmatter}



\section{Introduction}
\begin{figure}[t!]
  \centering

  \begin{subfigure}[t]{0.20\columnwidth}
    \includegraphics[width=\linewidth]{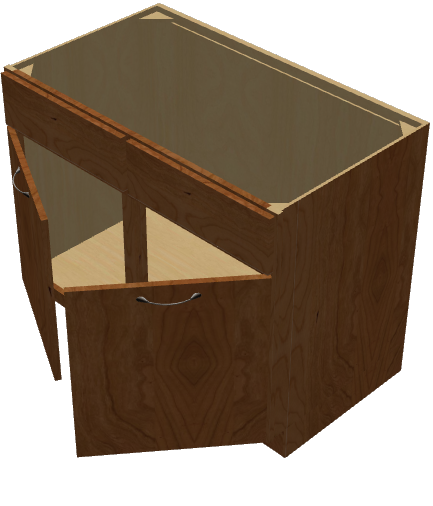}
    \caption{}
  \end{subfigure}%
  ~
  \begin{subfigure}[t]{0.70\columnwidth}
    \includegraphics[width=\linewidth]{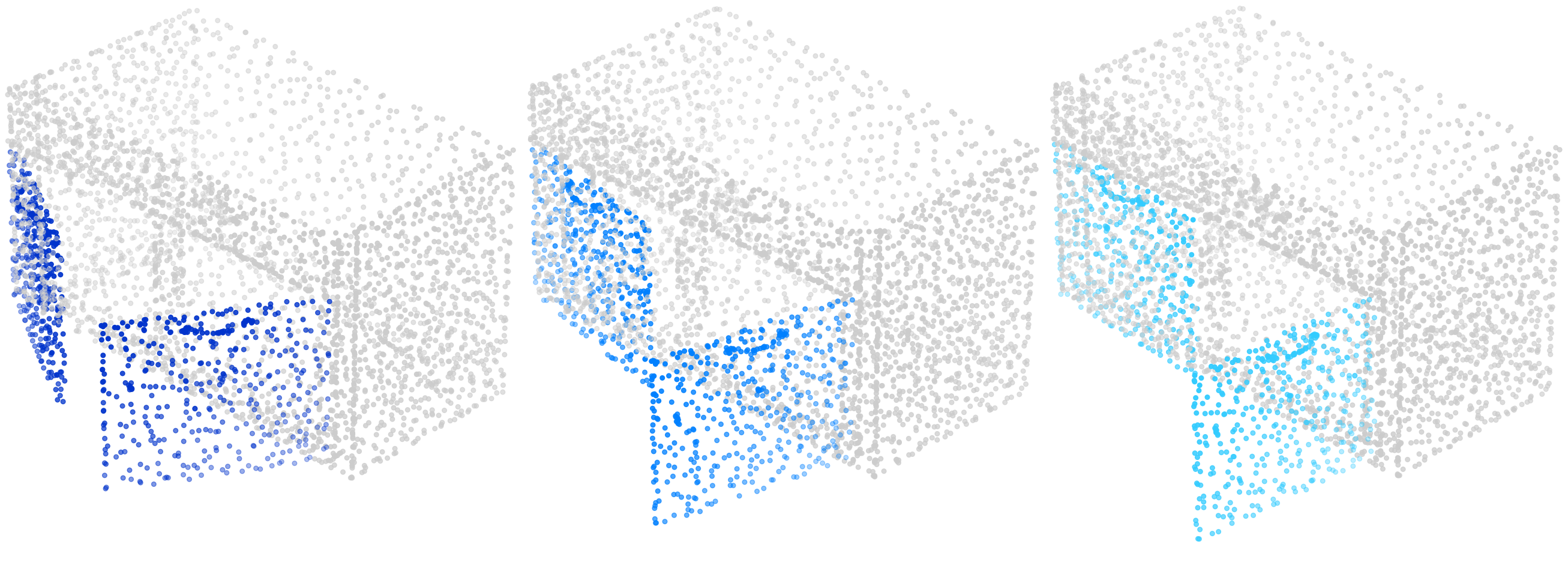}
    \caption{}
  \end{subfigure}

  \begin{subfigure}[t]{0.30\columnwidth}
    \adjincludegraphics[width=\linewidth,clip,trim={{0.50\width} {0.018\height} 0 {0.018\height}}]{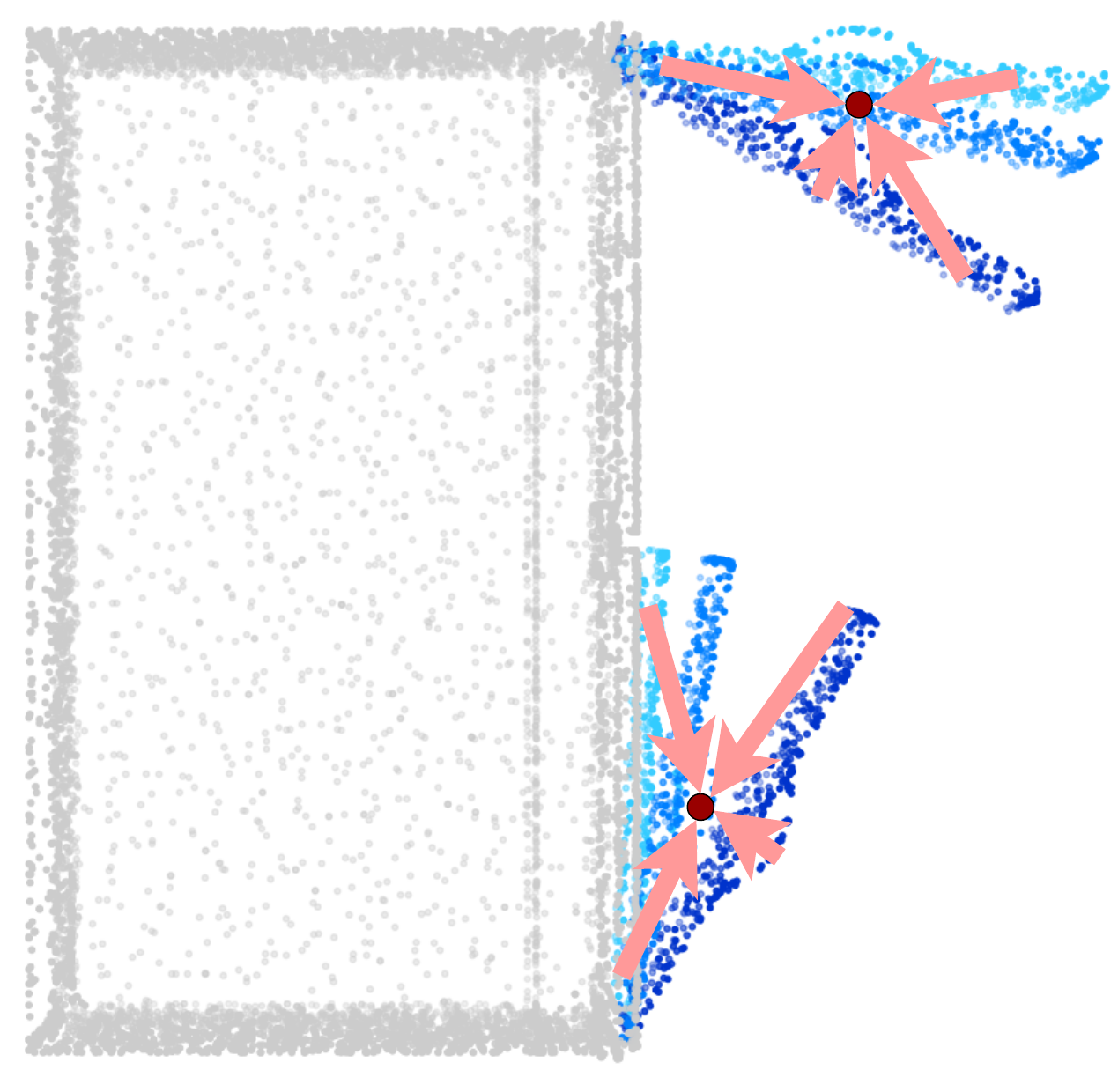}
    \caption{}
  \end{subfigure}%
  ~
  \begin{subfigure}[t]{0.30\columnwidth}
    \adjincludegraphics[width=\linewidth,clip,trim={{0.50\width} 0 0 0 }]{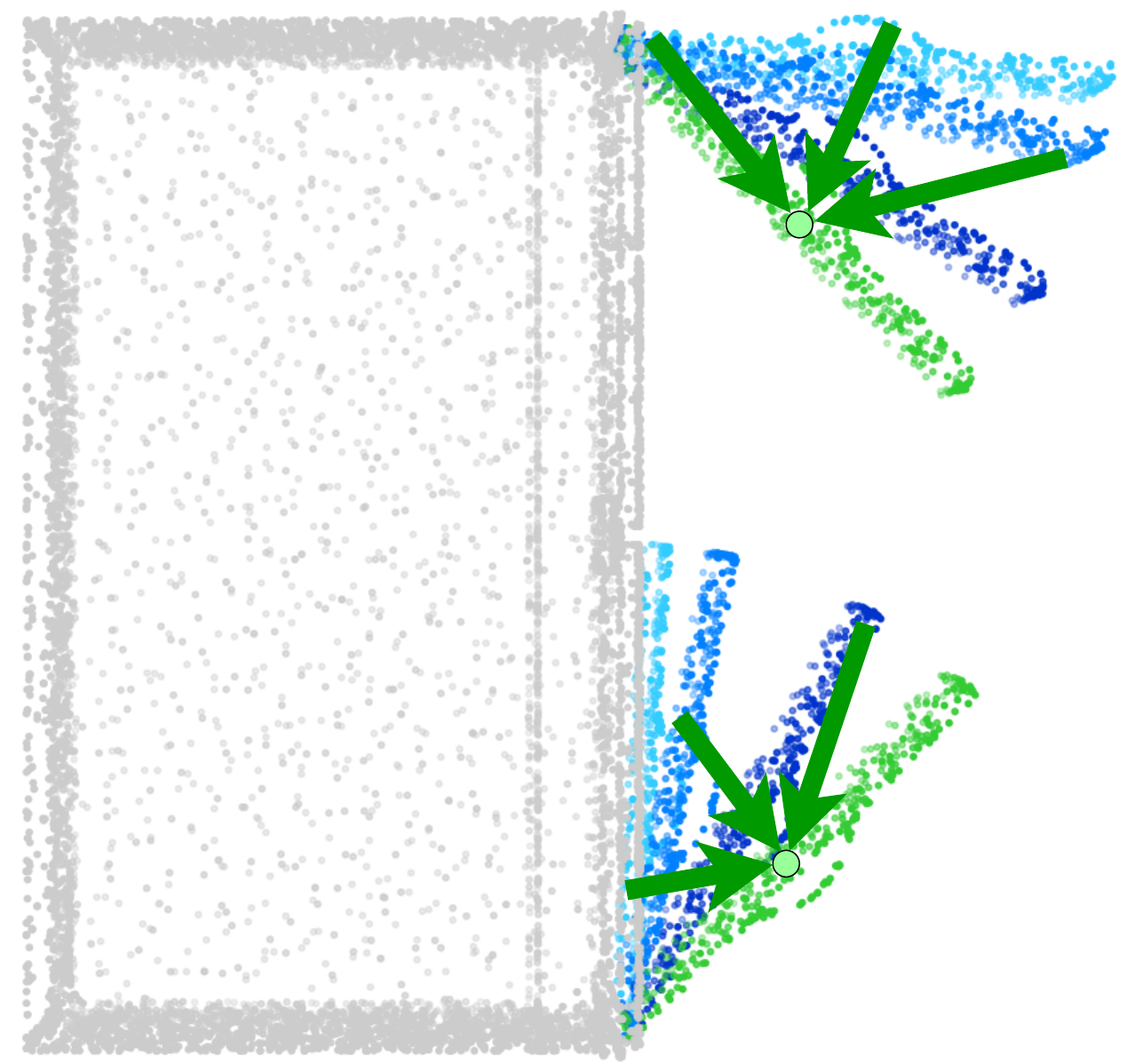}
    \caption{}
  \end{subfigure}

  \caption{
    4D panoptic segmentation for articulated objects.
    Given a 3D model of an articulated object (a), a 4D point cloud sequence is generated (b) by capturing the object in different articulation states.
    State-of-the-art methods (c) transform each point (red arrows) to the centroid of its 4D space-time part (red circle), which shifts with articulation, as seen when the centroid of the closed door is closer to the cabinet body than the one of the open door, resulting in inconsistent reference points.
    The proposed method (d) transforms each point into a learned canonical space (green part), yielding consistent representations (green arrows) across articulation states.
  }
  \label{fig:graphical-abstract}
\end{figure}
Object perception is a fundamental challenge in computer vision, crucial for robotics applications in scenarios where interaction with the environment is needed, such as service robotics, autonomous driving or healthcare robotics \cite{Tian2023,Ntakolia2023,Fernandez-Chaves2022}.
While significant research has focused on object detection \cite{Renfei2023,Chandrakar2022}, segmentation \cite{Chen2025, Jiang2024}, and pose estimation \cite{Zhang2026,Hoque2023}, most methods are designed for rigid objects, whose shapes do not change during manipulation.

However, many common objects are articulated, consisting of multiple parts connected by joints that allow for constrained relative motion, such as cabinets, drawers, and washing machines.
Perceiving articulated objects is particularly challenging as it requires understanding the dynamic relationships between their individual parts as they move \cite{Weng2024, Fu2024}.
Since this motion is inherently temporal, a major limitation of existing approaches is their insufficient use of temporal information to model these complex dynamics \cite{Zeng2024a, Heppert2023}.

A key, yet underexplored, challenge is the segmentation of articulated objects over time.
This task, known as 4D panoptic segmentation, involves assigning consistent semantic and instance labels to point cloud sequences across multiple frames \cite{Yilmaz2024, Zhu2023}.
Although 4D panoptic segmentation is well-studied in autonomous driving, methods from that domain do not readily generalize to articulated objects due to fundamental differences in scene structure, object scale, and motion patterns.
Progress in this area is further limited by the absence of a dedicated benchmark for articulated objects, as existing public datasets typically provide only 3D models rather than sensor data, hindering like-for-like comparison \cite{Liu2022a, Xiang2020, Mo2019a, Chang2015}.
To close this gap, we present Artic4D, a benchmark offering synchronized RGB images, depth maps, and colored point clouds of everyday articulated objects across diverse articulation states, with frame-level panoptic annotations and standardized splits to enable fair, reproducible comparison.

Building on this benchmark, we propose CanonSeg4D, a novel 4D panoptic segmentation method specifically designed for articulated objects.
This approach is based on a spatio-temporal encoder-decoder architecture that captures both the spatial and temporal features of the object.
A key novelty lies in how instance segmentation is performed.
Existing methods often transform points from the same instance to a common representation, such as the instance's centroid, to facilitate clustering.
However, for articulated objects, this centroid shifts with the object's articulation state, leading to an inconsistent representation.
To address this, the proposed method learns a canonical representation for each instance (i.e., each object part) that is independent of its articulation state.
As illustrated in Figure~\ref{fig:graphical-abstract}, this approach overcomes the limitations of state-of-the-art methods that rely on inconsistent reference points.
By transforming points into a stable canonical space, the proposed method decouples the instance's representation from its current pose, providing a more robust foundation for learning and leading to improved segmentation performance.

This paper makes three key contributions:
\begin{itemize}
  \item We introduce a novel dataset, Artic4D, which comprises 4D sensor data of articulated objects, designed to facilitate research in this domain;
  \item We propose a novel panoptic segmentation method, CanonSeg4D, for articulated objects. This method employs learned canonical representations to achieve segmentation that is invariant to the object's articulation state;
  \item We perform an extensive experimental evaluation of the proposed method on the Artic4D dataset, including a comparative analysis against state-of-the-art approaches, which demonstrates the superiority of CanonSeg4D in complex scenarios;
\end{itemize}
The source code implementing the proposed method, along with the dataset and its generation pipeline, will be released upon paper acceptance.
\section{Related Work}
\subsection{Articulated Object Perception}\label{sec:articulated-object-perception}
Articulated object perception is a complex field that involves various tasks such as reconstruction \cite{Song2024, Liu2023b}, pose estimation \cite{Liu2024b, Weng2021}, motion perception \cite{Liu2023, Chu2023}, and segmentation.
While this field has been growing, the manner in which these tasks are approached are still very limited, with the task of segmentation being totally not addressed.

A common technique is the exploitation of categorical models \cite{Liu2022,Li2020a}, which represent each object category with a single learned neural model.
Therefore, to perceive various types of objects, various models are needed.
These models achieve high accuracy but are limited to known categories, making them unsuitable for tasks where object categories may be unknown, like in robotic manipulation.
Another limitation is the use of multiple models simultaneously when analyzing a scene with objects from different categories.

A large number of methods rely on static data like 3D point clouds \cite{Liu2023, Chu2023} or images \cite{Zeng2024a, Heppert2023}.
This is a large downfall, as articulated objects are inherently dynamic.
Therefore, using only static data is neglecting the crucial temporal information needed to understand articulated movements.
While some works use temporal data, they are often limited to only two timestamps \cite{Weng2024, Jiang2022}, providing minimal temporal context.

\subsection{Synthetic Data}\label{sec:synthetic-data}

Collecting and annotating large-scale real-world datasets is both time-consuming and costly \cite{Song2024b}, especially for new tasks like articulated object segmentation.
These objects are challenging to collect since most methods require complete 3D point clouds from multiple viewpoints to resolve occlusions \cite{Jiang2022, Li2020a}.
This must be repeated for each articulation state and object type, which might drastically change in shape and size, resulting in a combinatorial explosion of sensor data to collect and annotate, often with different sensor setups and processing steps.

Synthetic data is an effective alternative to overcome these limitations \cite{Paulin2023, He2022, Nikolenko2021}, allowing for the generation of diverse data and annotations in a controlled environment.
This motivated the creation of PartNet Mobility \cite{Xiang2020, Mo2019a, Chang2015}, a synthetic dataset with over 2000 articulated 3D object models.
However, PartNet Mobility requires users to simulate motion and record sensor data themselves, which is time-consuming and can lead to inconsistencies, since the application is highly dependent on the approach.
Therefore, it is not a true benchmark dataset and does not allow for easy comparison between approaches.
Despite this, it has become the standard in the articulated object perception community.

\subsection{4D Panoptic Segmentation}\label{sec:4d-panoptic-segmentation}

4D panoptic segmentation uses spatio-temporal information to segment objects into semantic classes and instances consistently over time.
Most methods use 4D point clouds as input data.
A common approach is to estimate a transformation for each point to obtain a single representation per instance
After this transformation, a clustering algorithm is ran, separating points belonging to different instances.
This approach was pioneered in the 3D domain and was done by predicting offsets from each point to its instance centroid \cite{Jiang2020}.
4D methods extend this by using offsets towards the moving 4D instance's centroid, i.e., the spatial centroid of the 4D point cloud of each instance \cite{Zhu2023, Kreuzberg2023}.
Other methods use a mask-based approach, predicting a mask for each instance \cite{Yilmaz2024, Marcuzzi2023}.

Existing methods focus on autonomous driving scenarios.
Therefore, these methods make assumptions that hold valid for this scenario, such as working with rigid objects like cars and pedestrians \cite{Marcuzzi2022, Aygun2021}, all instances being on a ground plane \cite{Marcuzzi2023} or the point cloud being projectable to 2D \cite{Li2022a}.
This makes current approaches unsuitable for the the task of segmenting articulated objects.

Transformation-based methods assume a single offset to the 4D instance centroid is sufficient for segmentation \cite{Zhu2023, Kreuzberg2023}.
This is problematic for articulated objects, whose parts may move abruptly, unlike the smoother motions typical in autonomous driving
An offset computed from the instance's centroid can be misaligned with individual part positions, hindering accurate segmentation and part association across consecutive frames of the same sequence.
This inconsistency also complicates model training, as ground truth offsets vary significantly, making it difficult to learn robust representations.

\section{Artic4D Dataset}
Despite the availability of high-quality articulated CAD models in PartNet Mobility \cite{Xiang2020, Mo2019a, Chang2015}, fair and reproducible evaluation remains difficult as prior works often differ in how they sample articulation states, place cameras, render sensor data, and define labels, leading to incomparable experimental conditions.
This lack of standardization hinders meaningful benchmarking and slows progress in articulated object perception.

To address this gap, the Artic4D benchmark and its accompanying, reproducible generation pipeline are presented.
Artic4D provides synchronized RGB, depth, and point clouds of diverse articulated objects in various articulated states by rendering PartNet Mobility models in the SAPIEN physics engine \cite{Xiang2020}.
The dataset standardizes articulation trajectories, sensor viewpoints, and annotation conventions (instance- and semantic-level), enabling apples-to-apples comparisons across methods.
Equally important, the pipeline is parameterized and extensible: researchers can regenerate the dataset, add new objects, produce additional data modalities, and create new labels while preserving compatibility with the core benchmark.
This design supports both a fixed benchmark release and task-specific variants tailored to different research needs.

\subsection{Dataset Generation}

\begin{figure}[t]
  \centering
  \includegraphics[width=0.8\columnwidth]{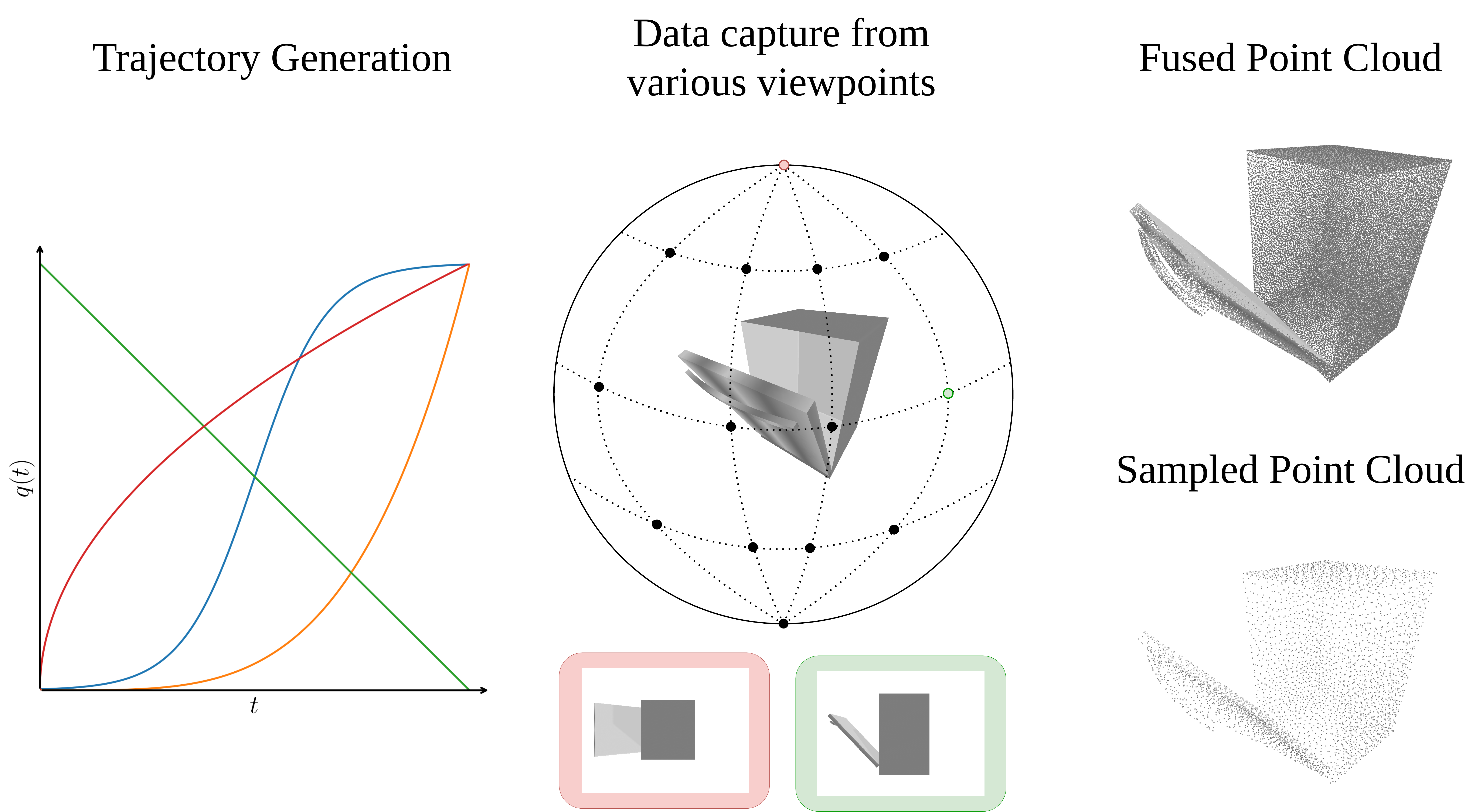}
  \caption{
    Artic4D dataset generation pipeline.
    Joint trajectories defined by power-law and sigmoid profiles (Eqs.~(\ref{eq:traj-pow}) and (\ref{eq:traj-sig})) are uniformly sampled at 100 articulation states (left panel).
    For each state, a set of RGB-D viewpoints uniformly distributed on a viewing sphere capture the object (middle panel).
    Depth maps from all viewpoints are fused into a single point cloud and downsampled via farthest point sampling (right panel).
  }
  \label{fig:dataset-pipeline}
\end{figure}

Figure~\ref{fig:dataset-pipeline} outlines the data generation pipeline.
For each object, articulation is simulated by generating per-joint trajectories, i.e., time profiles that map each timestamp to a joint configuration between its start and end limits.
A distinct trajectory is assigned to every actuated joint of every object.
To cover a range of plausible motion patterns, two temporal profiles are used.
These profiles model time, $t \in [0, T]$,  to a joint state $q$.

The first profile, $q_{\text{power}}$, is based on a power-law profile:
\begin{equation}
  q_{\text{power}}\mleft(t\mright) = q_0 + \left(q_f - q_0\right) \left(\frac{t}{T}\right)^a\,,
  \label{eq:traj-pow}
\end{equation}
where $q_0$ and $q_f$ are the start and end configurations.
The exponent $a$ dictates the motion profile.
An exponent greater than 1 results in an accelerating motion, while an exponent less than 1 causes a decelerating motion.
An exponent of 1 corresponds to a linear trajectory with constant velocity.

The second profile, $q_{\text{sig}}$ is based on a sigmoid:
\begin{equation}
  q_{\text{sig}}\mleft(t\mright) = q_0 + \left(q_f - q_0\right) \: \sigma\mleft(\frac{t}{T}\mright)\,, \quad \sigma(x)= \frac{1}{1+e^{6-12x}}\,,
  \label{eq:traj-sig}
\end{equation}
This profile provides a smooth "ease-in/ease-out" motion that mimics natural movements.
The part starts moving slowly, accelerates through the middle of its path, and then decelerates to a stop.

Both profiles admit a mirrored (inverse) trajectory, $q^{\text{inv}}$ that starts at $q_f$ and ends at $q_0$ via time reversal:
\begin{equation}
  q^{\text{inv}}\left(t\right) = q\left(T - t\right)\,,
\end{equation}
where $q = q_{\text{power}}$ for the power-law profile and $q = q_{\text{sig}}$ for the sigmoid profile
For each object, $n$ articulation states are uniformly sampled along the trajectory over $[0,T]$:
\begin{equation}
  t_i = i \,\frac{T}{n-1}\,, \quad q_i = q(t_i)\,, \quad i=0,\dots,n-1\,.
\end{equation}
where $i$ indexes the samples (with endpoints included), $t_i$ is the $i$-th time step, and $q_i$ is the joint state at time $t_i$.
Each articulation state can then be applied to the object, leading to a discrete but fluid motion of its parts.

To capture the sensor data, for each articulation state, a set $\mathcal{J}$ of RGB-D cameras are placed on an object-centric sphere of radius $r$.
Camera positions for viewpoint $j \in \mathcal{J}$, $\mathbf{v}_j$, are parameterized as:
\begin{equation}
  \mathbf{v}_j = r\left(\sin(\theta_j)\cos(\phi_j),\;\sin(\theta_j)\sin(\phi_j),\;\cos(\theta_j)\right)^\intercal
\end{equation}
where $\theta_j$ and $\phi_j$ are the altitude and azimuth angles for viewpoint $j$, respectively, and $(.)^\intercal$ is the transpose operator.
Altitude and azimuth angles for all viewpoints are chosen to approximate uniform coverage of the sphere, and all cameras are pointed to the center of the sphere, where the object is placed.
This surrounding coverage of the object allows for various types of data, from full articulation coverage to strong self-occlusion, allowing users to choose samples that best align with their goals, e.g., full-visibility baselines versus occlusion-stressed tests.
The cameras capture RGB images and depth maps.
Colored point clouds are then generated by combining RGB with depth.

Because the viewpoints surround the object, per-view point clouds are fused into a complete $360^\circ$ point cloud:
\begin{equation}
  P_i = \bigcup_{j=1}^{|\mathcal{J}|} P_{i,j}\,, \qquad \tilde{P}_i = f\mleft(P_i, m\mright)\,,
\end{equation}
where $P_{i,j}$ denotes the point cloud at articulation state $i$ from viewpoint $j$, $P_i$ is the fused $360^\circ$ point cloud for state $i$,
$|\cdot|$ denotes set cardinality, $\tilde{P}_i$ is the fused cloud downsampled to $m$ points, and $f(\cdot, m)$ denotes farthest point sampling, with $m$ the target size.
Downsampling controls the otherwise very large number of points produced by the fusion.

All parameters and random seeds are fixed and released with the pipeline, making generation deterministic and fully regenerable.
This standardization ensures fair comparisons while remaining easily extensible to new objects, modalities, and labels.

\subsection{Dataset Description}
\begin{table}[t]
  \centering
  \caption{Artic4D dataset statistics by category.}
  \begin{tabular}{lrlr}
    \toprule
    Category   & Models & Category     & Models \\
    \midrule
    Storage    & 346    & Table        & 101 \\
    Eyeglasses & 65     & USB          & 51 \\
    Laptop     & 48     & Scissors     & 46 \\
    Knife      & 45     & Refrigerator & 43 \\
    Dishwasher & 41     & Safe         & 30 \\
    Door       & 29     & Window       & 28 \\
    Oven       & 24     & Pliers       & 24 \\
    Washing M. & 17     & Microwave    & 13 \\
    \midrule
    \multicolumn{3}{r}{Total Models:}  & 951 \\
    \bottomrule
  \end{tabular}
  \label{tab:dataset-stats}
\end{table}

Table~\ref{tab:dataset-stats} summarizes model counts by category.
The benchmark covers 16 categories and 951 CAD models.
For each model, 100 articulation states are sampled and 18 viewpoints per state are captured ($18\times 100=1800$ viewpoints per model), for a total of 1,711,800 viewpoints.
Each viewpoint includes three modalities: an RGB image, a depth map, and a colored point cloud, totaling 5,135,400 samples.
The dataset also provides one fused $360^\circ$ colored point cloud per articulation state and model (100 per model), adding 95,100 fused point clouds for view-agnostic evaluation.
Overall, the release comprises 5,230,500 data samples.
The chosen categories follow prior work \cite{Zeng2024a,Weng2024,Chu2023,Heppert2023}, the imbalance in model counts mirrors the availability of PartNet Mobility assets within these categories.

Annotations include articulation parameters (joint type, axis, anchor point, motion limits) and both instance- and semantic-level segmentation.
Instance labels index distinct articulated parts of an object.
Semantic labels describe part function independent of object category.
Six semantic classes are defined by movement affordances: Body, Drawer, Hinged Door, Lid, Leg, and Slider.
Body denotes the main, typically fixed, structure.
Drawer denotes the part that translates in an in-out motion, such as desk drawers.
Hinged Door denotes the front panel that rotates to close an opening; hinges may be vertical (side-hinged, e.g., cabinets) or horizontal along the bottom edge (down-hinged, e.g., ovens, dishwashers).
Lid denotes the top-cover panel that rotates about a rear hinge to close a top-facing opening, such as laptop screens, washing-machine lids.
Leg denotes rotational appendages, such as eyeglass temples, scissor and plier handles.
Slider denotes a panel that translate along a track, such as window panes, sliding doors.

\newcolumntype{C}{ >{\centering\arraybackslash} m{2.5cm} }
\newcolumntype{D}{ >{\centering\arraybackslash} m{2cm} }

\begin{table*}[t!]
  \centering
  \caption{
    Examples of ground-truth semantic labels in Artic4D.
    Each row shows two point-cloud renderings from different object categories that share the same semantic class, highlighting category-agnostic labeling.
    Colors denote classes: Body (blue), Drawer (red), Hinged Door (orange), Lid (green), Leg (brown), Slider (purple).
    The rightmost image shared by the Drawer and Hinged Door rows illustrates co-occurrence of both classes on a single object.
  }
  \begin{tabular}{D C C}
    \toprule
    Semantic & \multicolumn{2}{c}{\multirow{2}{*}{Examples}} \\
    Class & & \\
    \midrule
    Drawer         & \includegraphics[height=2cm]{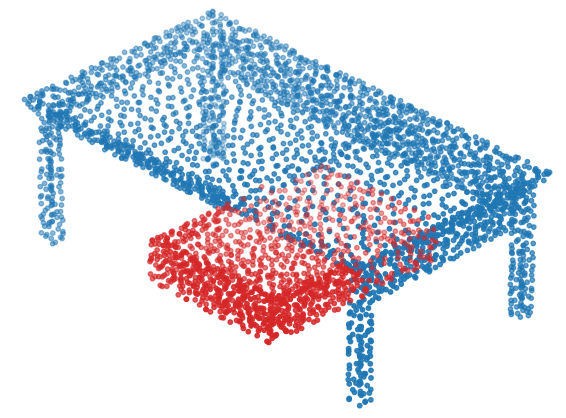} & \multirow{2}{*}{\includegraphics[height=2cm]{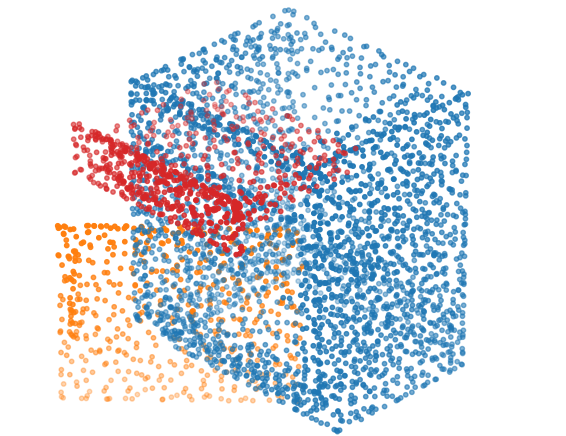}} \\
    Hinged Door    & \includegraphics[height=2cm]{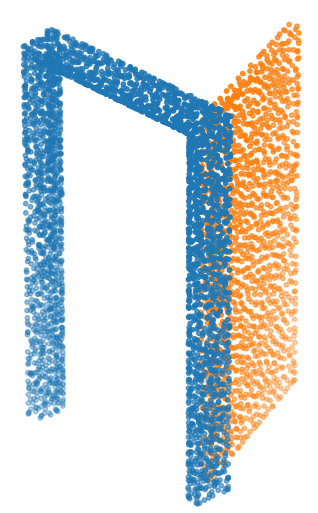}   &                                                                                     \\
    Lid            & \includegraphics[height=2cm]{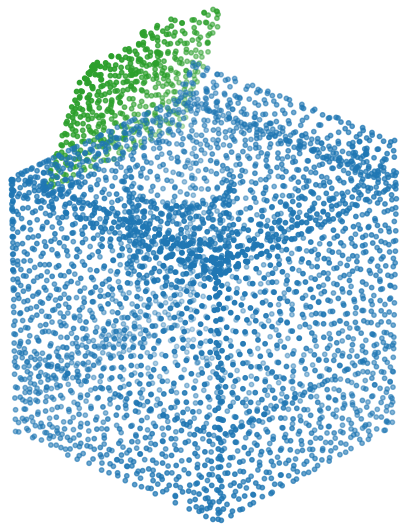}    & \includegraphics[height=2cm]{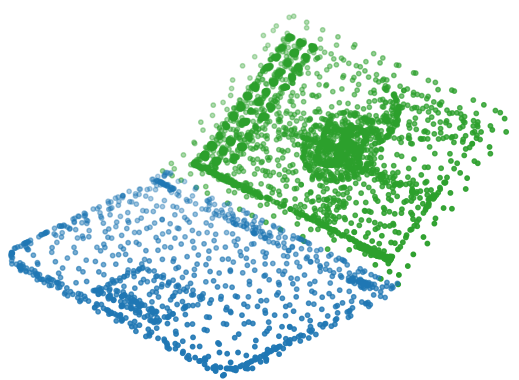}                        \\
    Leg            & \includegraphics[height=1.33cm]{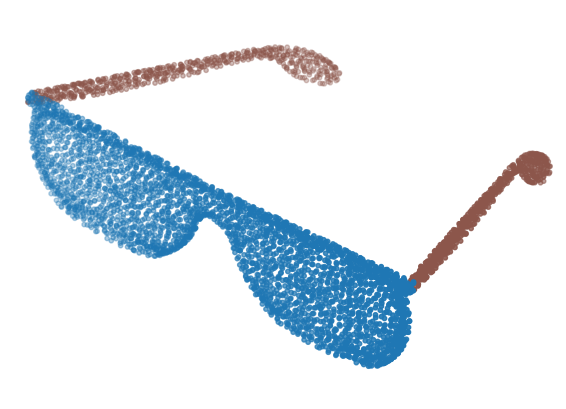} & \includegraphics[height=1.33cm]{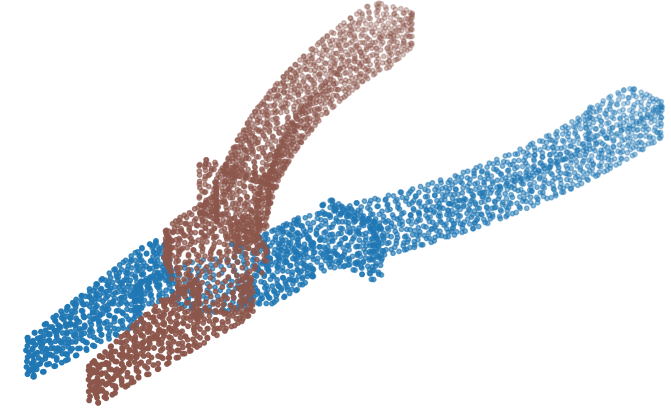}                     \\
    Slider         & \includegraphics[height=2cm]{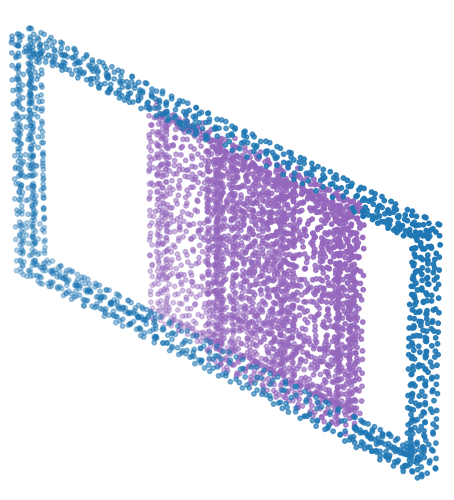} & \includegraphics[height=2cm]{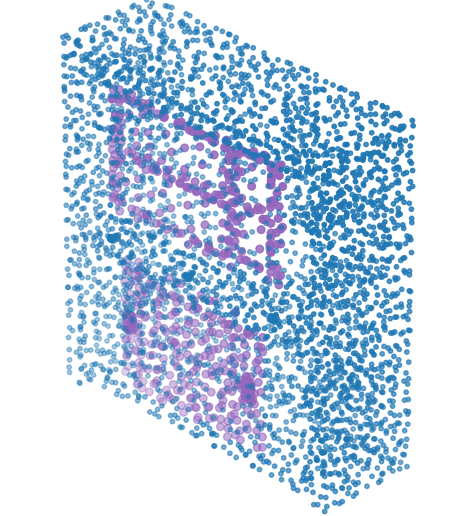}                     \\
    \bottomrule
  \end{tabular}
  \label{tab:semantic-classes}
\end{table*}

Table~\ref{tab:semantic-classes} illustrates that semantic classes are category-agnostic and can co-occur within a single object.
All objects include the Body class, which is typically the majority.
Categories may contain multiple semantic classes, e.g. Storage can include Body, Drawer, Hinged Door, and Slider, and a single sample can contain several classes simultaneously, e.g., Body, Drawer, and Hinged Door.

Given the extensive nature of the Artic4D dataset with its vast number of samples and varied articulation complexities, it was strategically divided into three subsets to address evaluation challenges across different difficulty levels.
Artic4D-S contains objects with only a single moving part, establishing a baseline difficulty.
Artic4D-D includes more complex objects with two moving parts, presenting greater challenges.
Artic4D-M combines both categories, offering a more varied bundle of different articulation types.
Since this division is based on articulation complexity rather than object category, none of the individual subsets span the complete range of 16 object categories present in the full dataset.
This structured division allows researchers to obtain more comprehensive results by systematically assessing method performance across different difficulty levels, facilitating nuanced comparisons and standardized benchmarking of articulated object understanding approaches.

\section{Method}

\subsection{Problem Definition}
Given a sequence of 4D point clouds representing an articulated object, let $X \in \mathbb{R}^{S \times N \times \left(3 + D_i\right)}$ denote the input, where $S$ is the number of time frames in the sequence, $N$ is the number of points in each frame, and each point is described by its 3D coordinates plus $D_i$ additional features, such as RGB color or LiDAR intensity.
The objective of 4D panoptic segmentation is to jointly perform semantic and instance segmentation for each point in every frame.
Semantic segmentation assigns each point a label from a set of $k$ semantic classes, $\mathcal{C} = \{c_1, \ldots, c_k\}$.
Instance segmentation groups points into one of $m$ distinct object parts, $\mathcal{P} = \{p_1, \ldots, p_m\}$, ensuring that instance identities remain consistent across all frames in the sequence.

\subsection{Architecture}

\begin{figure*}[t!]
  \centering
  \includegraphics[width=0.85\textwidth]{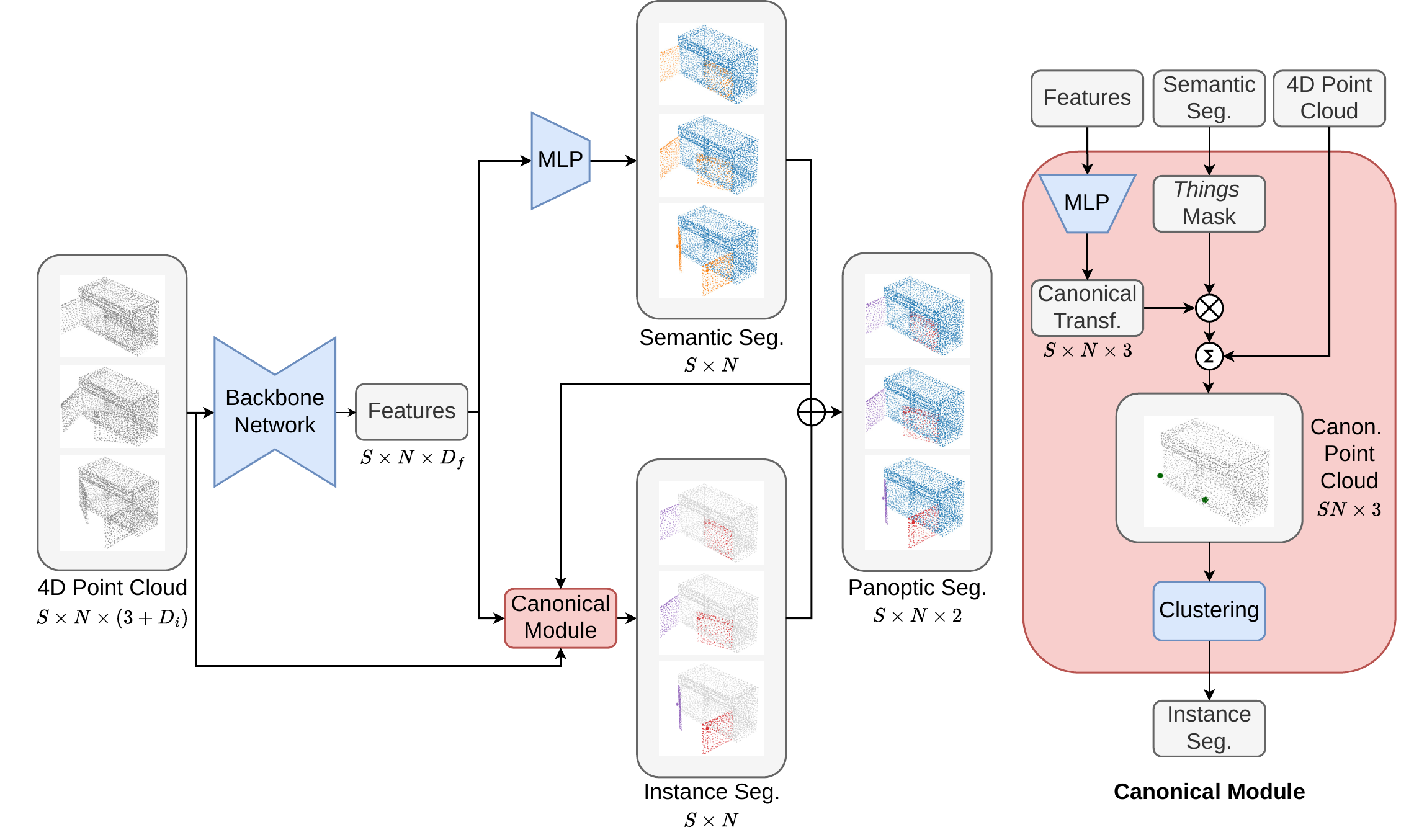}
  \caption{Overview of CanonSeg4D, the proposed 4D panoptic segmentation architecture.
    The input is a 4D point cloud, which is processed by a feature extraction backbone to capture both spatial and temporal features.
    From those features, a segmentation head predicts semantic labels and the canonical module predicts instance labels.
    Both outputs are combined to achieve panoptic segmentation.
    The canonical module, in red, learns to transform each point into a canonical space representation, creating well-defined part centroids.
    These centroids are then used to group points into instances, using a clustering algorithm.
  }
  \label{fig:architecture}
\end{figure*}

Figure~\ref{fig:architecture} illustrates the architecture of CanonSeg4D.
The proposed architecture consists of three main components: a feature extraction backbone, a semantic head, and a canonical module for instance identification.
The following sections describe each component in detail.

\subsubsection{Feature Extraction Backbone}

The feature extraction backbone needs to extract rich and meaningful features from the input 4D point cloud sequence.
Therefore, PST-Transformer \cite{Fan2023} was chosen.
PST-Transformer is a state-of-the-art architecture that effectively captures both spatial features and temporal dynamics across frames.
This model operates by first creating a sampling mechanism where each sampled point establishes a radius of influence for learning local spatial features.
This influence radius extends temporally across the entire sequence, forming what is called a "point tube", which enables the model to capture temporal dependencies between frames.
The sampling approach significantly reduces the point cloud density, creating a concise feature map similar to PointNet++ \cite{Qi2017a}.
These sampled points then pass through a transformer architecture, which leverages attention mechanisms to learn meaningful relationships between local features.
Finally, the model decodes these learned representations back to the original 4D point cloud dimensions, producing a feature vector for each point that encodes both spatial structure and temporal dynamics.
The backbone's ability to gracefully handle both spatial and temporal information, while effectively extracting both local and global features, were the key reasons for its selection in this architecture.

\subsubsection{Semantic Head}

The semantic head is responsible for predicting semantic labels for each point in the 4D point cloud sequence.
It takes the feature vectors produced by the backbone and applies an \ac{MLP} to output a semantic label for each point.

The segmentation head is trained with the Lovász-Softmax loss~\cite{Berman2018}, $\mathcal{L}_{\text{sem}}$, which directly optimizes a differentiable surrogate of the mean \ac{IoU} metric.
A key motivation for using the Lovász-Softmax loss in this context is the inherent class imbalance present in articulated objects, as most articulated objects have a large majority of their volume occupied by the "body", a fixed, non-articulated part, while the articulated parts are typically much smaller.
By optimizing for mean \ac{IoU}, which weighs each class equally regardless of frequency, Lovász-Softmax mitigates the dominance of large classes and improves segmentation of smaller, articulated parts.

\subsubsection{Canonical Module}
The canonical module transforms points from the 4D point cloud sequence into a learned canonical space.
Canonical spaces are standardized reference frames commonly used in articulated object pose estimation \cite{Liu2022, Li2020a}.
In pose estimation, they normalize an object's scale, position, and articulation state to a predefined configuration.
In this work, this concept is extended to 4D panoptic segmentation by mapping each object part to a consistent, learned representation, regardless of its changing position or articulation across frames.
The objective is to transform all points of each part instance onto a single point, making different points for each instance.
The transformation is a two-step process: first, mapping the points to a canonical articulation state, and second, mapping those points to their respective centroid.

This transformation is based on the principles of forward kinematics.
The pose of part $p$ with respect to the origin frame $o$ is given by the homogeneous transformation matrix ${}^oT_{p}\left(q\right)$, a function of the joint configuration vector $q$.
It is obtained as the ordered product of the joint transformations along the kinematic chain from $o$ to $p$:
\begin{equation}
  {}^oT_{p}\left(q\right) = \prod_{i=1}^{n_p} A_i\left(q_i\right)\,,
  \label{eq:transform}
\end{equation}
\noindent where $n_p$ is the number of joints on that chain and $A_i(q_i)$ is the homogeneous transformation for joint $i$ with joint variable $q_i$.

Using this, a point $\mathbf{x}_p^{\left(t\right)}$ on part $p$ at time $t$ (with joint configuration $q_t$) is mapped to its corresponding position in a predefined canonical joint configuration $q_c$.
This is achieved by first applying the inverse transformation at time $t$ to bring the point to its local part frame, and then applying the forward transformation for the canonical configuration:
\begin{equation}
  \mathbf{x}_{p, \text{canon}} = {}^oT_{p}\left(q_c\right) \left({}^oT_{p}\left(q_t\right)\right)^{-1} \mathbf{x}_p^{\left(t\right)}\,,
  \label{eq:x_canon}
\end{equation}
\noindent where $\mathbf{x}_{p, \text{canon}}$ is the position of the point $p$ in the canonical articulation state.

Finally, to create a single, consistent target for all points within a part, each point is mapped to the centroid of that part's canonical representation.
This collapses all points of a part into a single point in space, which serves as the canonical instance center:
\begin{equation}
  \mathbf{x}_{p, \text{target}} = \frac{1}{|\mathcal{X}_{p, \text{canon}}|} \sum_{\mathbf{x} \in \mathcal{X}_{p, \text{canon}}} \mathbf{x} = c_{p, \text{canon}}\,,
  \label{eq:x_centered}
\end{equation}
\noindent where $\mathbf{x}_{p, \text{target}}$ is the final target position for all points on part $p$, $\mathcal{X}_{p, \text{canon}}$ is the set of all points belonging to part $p$ in the canonical state, which was obtained by transforming all points using equation (\ref{eq:x_canon}), and $c_{p, \text{canon}}$ is their centroid.

The ground truth transformation for the point $\mathbf{x}_p^{\left(t\right)}$ is then simply the offset vector, $\mathbf{o}_p^{\left(t\right)}$, pointing from its original position to this target:
\begin{equation}
  \mathbf{o}_p^{\left(t\right)} = \mathbf{x}_{p, \text{target}} - \mathbf{x}_p^{\left(t\right)}\,.
\end{equation}
\noindent This transformation is only applied to points belonging to \textit{things} classes (movable parts).
Points from \textit{stuff} classes (static or background elements) are not transformed, which is consistent with standard 4D panoptic segmentation practices \cite{Zhu2023, Aygun2021}.

The transformation is estimated through an \ac{MLP} that processes feature vectors from the backbone network.
Training this component involves a dual-term loss function.
First, an L1 loss that minimizes the L1 distance between the predicted and ground truth offset vectors:
\begin{equation}
  \mathcal{L}_{\text{dist}}\left(\hat{\mathbf{o}}, \mathbf{o}\right) = \|\hat{\mathbf{o}} - \mathbf{o}\|_1\,,
\end{equation}
\noindent where $\|\cdot\|_1$ denotes the L1 norm, $\hat{\mathbf{o}}$ is the predicted offset vector, and $\mathbf{o}$ is the ground truth offset vector.

Second, an angle loss using cosine similarity enforces directional alignment between the predicted and ground truth offsets:
\begin{equation}
  \mathcal{L}_{\text{angle}}\left(\hat{\mathbf{o}}, \mathbf{o}\right) = 1 - \cos\alpha
  = 1 - \frac{\hat{\mathbf{o}} \cdot \mathbf{o}}{\|\hat{\mathbf{o}}\|_2 \, \|\mathbf{o}\|_2}\,.
\end{equation}
where $\alpha$ is the angle between $\hat{\mathbf{o}}$ and $\mathbf{o}$, $\|\cdot\|_2$ denotes the Euclidean norm, and $\cdot$ denotes the dot product.

The total loss for the canonical module is then:
\begin{equation}
  \mathcal{L}_{\text{canon}} = \frac{1}{\mathcal{S}\mathcal{N}} \sum_{s=1}^{\mathcal{S}} \sum_{n=1}^{\mathcal{N}} \left(  m_n^{\left(s\right)}
    \left(\mathcal{L}_{\text{dist}}\left(\hat{\mathbf{o}}_n^{\left(s\right)}, \mathbf{o}_n^{\left(s\right)}\right)
    + \mathcal{L}_{\text{angle}}\left(\hat{\mathbf{o}}_n^{\left(s\right)}, \mathbf{o}_n^{\left(s\right)}\right)\right)
  \right)\,,
\end{equation}

\noindent where $s$ indexes the frames in the sequence set $\mathcal{S}$, $n$ indexes the points in the set $\mathcal{N}$, $\hat{\mathbf{o}}_n^{\left(s\right)}$ and $\mathbf{o}_n^{\left(s\right)}$ are the predicted and ground truth offset vectors for point $n$ in frame $s$, respectively, and $m_n^{(s)}$ is a binary mask indicating whether point $n$ in frame $s$ belongs to a \textit{things} class.

After obtaining the canonical space representations of the points, a clustering algorithm is applied to group points into part instances.
Since the canonical space representation provides a consistent reference frame for each part, clustering can effectively identify distinct instances of articulated parts across frames.
Specifically, this work employs a density-based clustering algorithm which groups points according to their spatial proximity in the canonical space, without requiring prior knowledge of the number of instances.
This approach is particularly effective for handling varying numbers of parts and complex articulations, as it can adaptively discover clusters corresponding to individual part instances.
The output of the clustering process is a set of part instance masks, where each instance is assigned a unique identifier that remains consistent throughout the sequence.

\subsubsection{Optimization and Loss Propagation}
The network is trained end-to-end using the total loss
\begin{equation}
  \mathcal{L} = \mathcal{L}_{\text{sem}} + \mathcal{L}_{\text{canon}}\,.
\end{equation}
The feature extraction backbone is updated using gradients from both loss terms, ensuring that it learns representations beneficial for both semantic and canonical tasks.
In contrast, the MLPs in the semantic head and canonical module are each updated only by their respective loss functions, allowing them to specialize in their designated tasks.

\section{Experiments}
\subsection{Implementation Details}
All experiments were conducted on a single NVIDIA RTX 4090 GPU (24 GB).
We used the Artic4D dataset and its subsets, taking the provided colored point clouds as input.
Each 4D sample was constructed by concatenating three frames of colored point clouds into a sequence, where each point is represented by a 6-dimensional vector: its 3D spatial coordinates $(x, y, z)$ and RGB color values $(r, g, b)$.
To stress-test temporal robustness, the three frames in each sequence were chosen to be maximally spaced in time within the available clip, creating worst-case inter-frame motion.

The PST-Transformer architecture consists of a convolutional encoder-decoder backbone with an intermediate transformer module that processes the encoded feature representations.
Point cloud sampling is performed prior to encoding, followed by local feature aggregation for each sampled point.
The local neighborhood is defined as the 32 nearest neighbors within a spatial radius of 0.9 meters.
The aggregated neighborhood features are subsequently processed by the convolutional encoder.
The encoder architecture comprises four convolutional layers with progressive downsampling.
The first three layers apply a downsampling factor of 4, while the fourth layer applies a downsampling factor of 2.
The feature channels for these layers are 128, 256, 512, and 1024, respectively.
Temporal convolution with a kernel size of 3 is applied exclusively at the third layer, while all other layers employ a temporal kernel size of 1.
Each convolutional layer is followed by batch normalization and ReLU activation.
The transformer module, situated between the encoder and decoder, consists of 2 layers with 4 attention heads and a hidden dimension of 256.
The decoder architecture mirrors the encoder structure, employing four deconvolutional layers that progressively upsample the feature representations to reconstruct the original point cloud resolution with $D_f = 128$ features per point.
For each articulated part, the canonical state is defined as a reference configuration where the part is positioned at the midpoint of its motion range, i.e., halfway between its minimum and maximum positions.
This provides a consistent, motion-independent representation that serves as the transformation target for all instances of the same part type across different objects.

The model was trained using the AdamW optimizer \cite{Loshchilov2018} with an initial learning rate of $1 \times 10^{-3}$ and $\beta_1 = 0.9$, $\beta_2 = 0.999$, and weight decay of 0.01.
A warmup phase of 10 epochs was used, after which the learning rate was halved if no improvement was observed for 10 consecutive epochs.
To ensure reproducibility, all stochastic operations, including point cloud sampling and model parameter initialization, are seeded with a fixed random seed value of 42.

\subsection{Evaluation Metrics}
We use the standard metrics for 4D panoptic segmentation \cite{Aygun2021}.
The original notation is slightly adjusted to avoid variable conflicts.
These metrics evaluate the quality of the segmentation results by measuring both semantic and instance segmentation performance.
The first metric, $S_{cls}$, measures the semantic segmentation accuracy by computing the mean \ac{IoU} for all semantic classes:
\begin{equation}
  S_{cls} = \frac{1}{|C|}\sum_{c \in C} IoU\mleft(c\mright)\,.
\end{equation}
\noindent The second metric, $S_{assoc}$, evaluates the instance segmentation quality by measuring the consistency of instance associations across frames:
\begin{equation}
  S_{assoc} = \frac{1}{|\mathcal{B}|}\sum_{b \in \mathcal{B}} \frac{1}{gt_{id}\mleft(b\mright)} \sum_{\substack{a \in \mathcal{A} \\ a \cap b \neq 0}} TPA\mleft(a,b\mright) \, IoU\mleft(a,b\mright)\,,
\end{equation}
\noindent where $b$ is one instance in the set of ground truth instances $\mathcal{B}$, $gt_{id}\mleft(b\mright)$ represents the ground truth set of points belonging to instance $b$, $a$ is one instance in the set of predicted instances $\mathcal{A}$, and $TPA\mleft(a,b\mright)$ is the true positive association between predicted instance $a$ and ground truth instance $b$.
The last metrics is a combination between the semantic and instance segmentation metrics, which is defined as:
\begin{equation}
  LSTQ = \sqrt{S_{cls} \times S_{assoc}}\,.
\end{equation}

\subsection{Comparisons with the State-of-the-Art Methods}

For comparative analysis, several state-of-the-art methods were selected based on their leading performance on the 4D Panoptic Segmentation SemanticKITTI benchmark leaderboard\footnote{\href{https://semantic-kitti.org/tasks.html\#panseg4d}{https://semantic-kitti.org/tasks.html\#panseg4d}}.
The chosen methods represent top-ranked approaches in 4D panoptic segmentation, including three of the top four and the first-place model, Mask4Former.
The only top-four method excluded from these experiments was Mask4D \cite{Marcuzzi2023}, because it assumes that instances remain on the ground plane, which does not hold in this scenario.

\textbf{4D-StOP}~\cite{Kreuzberg2023} is a transformation-based method that translates each point to the centroid of its corresponding 4D instance.
The method also computes per-point geometric features to improve the clustering of points within the same instance.

\textbf{Eq-4D-StOP}~\cite{Zhu2023} extends 4D-StOP by incorporating principles of geometric deep learning.
It employs equivariant and invariant features to leverage underlying data symmetries, thereby enhancing segmentation accuracy.

\textbf{Mask4Former}~\cite{Yilmaz2024} utilizes a transformer-based architecture to learn a set of queries that encode both semantic and geometric information from the point cloud.
These queries are subsequently processed by a mask module to generate instance and semantic segmentation masks.

\begingroup
\setlength{\tabcolsep}{1.5pt}
\begin{table*}[t!]
  \centering
  \caption{
    Quantitative results of CanonSeg4D and state-of-the-art methods, 4D-StOP~\protect\cite{Kreuzberg2023}, Eq-4D-StOP~\protect\cite{Zhu2023}, and Mask4Former~\protect\cite{Yilmaz2024}, on the Artic4D dataset subsets.
    The metrics are $S_{cls}$, $S_{assoc}$, and $LSTQ$, which measure semantic segmentation, instance segmentation, and their combination, respectively.
    Results are reported in percentage, with larger values indicating better performance.
    Best results are highlighted in bold, second best are underlined.
  }
  \begin{tabular}{lccccccccc}
    \toprule
    &                 \multicolumn{3}{c}{Artic4D-S}                                 & \multicolumn{3}{c}{Artic4D-D}                             & \multicolumn{3}{c}{Artic4D-M}                              \\
    \cmidrule{2-10}
    &                 $S_{cls}$             & $S_{assoc}$       & $LSTQ$            & $S_{cls}$         & $S_{assoc}$       & $LSTQ$            & $S_{cls}$         & $S_{assoc}$       & $LSTQ$             \\
    \midrule
    4D-StOP             & 20.62             & 37.92             & 27.96             & 39.56             & 26.47             & 32.36             & 21.78             & 25.19             & 23.42              \\
    Eq-4D-StOP          & 19.40             & 41.10             & 28.24             & 41.30             & 28.40             & 34.29             & 23.00             & 25.28             & 24.11              \\
    Mask4Former         & \underline{56.20} & \textbf{81.17}    & \underline{67.56} & \underline{42.73} & \textbf{75.65}    & \underline{56.85} & \underline{58.95} & \underline{80.10} & \underline{68.71}  \\
    CanonSeg4D    & \textbf{87.13}    & \underline{81.07} & \textbf{84.05}    & \textbf{79.64}    & \underline{69.15} & \textbf{74.21}    & \textbf{88.94}    & \textbf{83.90}    & \textbf{86.39}     \\
    \bottomrule

  \end{tabular}
  \label{tab:quantitative-results}
\end{table*}
\endgroup

Table~\ref{tab:quantitative-results} presents the quantitative results of CanonSeg4D, our method, and the state of the art methods.
The results show that the proposed method achieves the best overall performance, particularly excelling in the most challenging, highly articulated scenarios, as reflected by the $LSTQ$ metric.

4D-StOP and Eq-4D-StOP demonstrate similar performance patterns across all datasets, with Eq-4D-StOP showing marginal improvements.
This slight advantage likely stems from its exploitation of geometric symmetries present in articulated objects.
However, both methods exhibit relatively poor performance overall.
On Artic4D-S, despite the simpler scenario with single articulated parts, these methods achieve limited success in instance segmentation, with Eq-4D-StOP reaching only 41.10\% for $S_{assoc}$.
Their semantic classification performance is particularly weak ($\leq$ 20.62\%), likely due to the inherent class imbalance in articulated object datasets, where body parts typically dominate the point distribution.
For Artic4D-D, an interesting reversal occurs - classification scores improve while association scores decline.
The improvement in semantic segmentation (up to 41.30\%) may be attributed to the increased diversity of articulated parts, providing more discriminative features for semantic classification.
Conversely, the decrease in association performance reflects the additional complexity introduced by multiple moving parts.
On the most challenging Artic4D-M subset, both semantic and instance metrics deteriorate significantly, indicating these methods struggle with complex scenarios featuring multiple articulated parts with diverse motion patterns.
This comprehensive performance decline suggests fundamental limitations in handling both the semantic diversity and motion complexity present in highly articulated objects.

Mask4Former shows poor semantic classification, likely due to the naturally imbalanced class distribution in articulated objects.
Their use of cross-entropy loss further amplifies this imbalance, leading to underrepresentation of minority classes.
Additionally, Mask4Former predicts both instance and semantic labels from the same learned queries, which works in autonomous driving scenarios but fails here.
A possible reason is that these queries capture local features better than global ones.
In articulated objects, especially those with rotational joints, the moving part may show very little overlap between frames, so relying primarily on local features is less effective.
Despite this, Mask4Former achieves strong instance segmentation.
It can be argued that this is because instance grouping relies mainly on geometric and spatial continuity, which remains robust even with complex articulation.
Physical separation between parts is often preserved, allowing the model to delineate instances accurately, even if semantic labels are incorrect.
Thus, instance segmentation metrics can remain high even when semantic segmentation is limited.

Our approach, CanonSeg4D, demonstrates consistently strong performance, ranking either first or second across all evaluation metrics.
Its results are particularly impressive in the most challenging scenarios on the Artic4D-M dataset.
This superior performance can be attributed to the method's canonical transformation approach, which establishes consistent reference configurations across different object types.
This consistency enables the model to generalize effectively across diverse articulated objects and even transfer knowledge between different object categories.
CanonSeg4D significantly outperforms both 4D-StOP and Eq-4D-StOP across all datasets, with performance gaps of 40-60 percentage points in many metrics.
This substantial difference highlights these methods' limited adaptability when applied to articulated object segmentation, a domain markedly different from autonomous driving scenarios for which these approaches were originally designed.
When compared to Mask4Former, CanonSeg4D shows substantially improved semantic segmentation capabilities.
This improvement stems from a specialized semantic loss function that effectively addresses the class imbalance inherent in articulated object datasets.
Additionally, our approach's dedicated 4D point cloud feature extraction backbone excels at capturing both local geometric details and global spatio-temporal relationships in the data.
By achieving a better balance between semantic classification ($S_{cls}$) and instance association ($S_{assoc}$), CanonSeg4D delivers higher overall $LSTQ$ scores.
This balanced performance profile becomes increasingly advantageous as articulation complexity increases, explaining the method's particular effectiveness in the most challenging scenarios.

\begin{figure}[t!]
  \centering
  \includegraphics[width=\columnwidth]{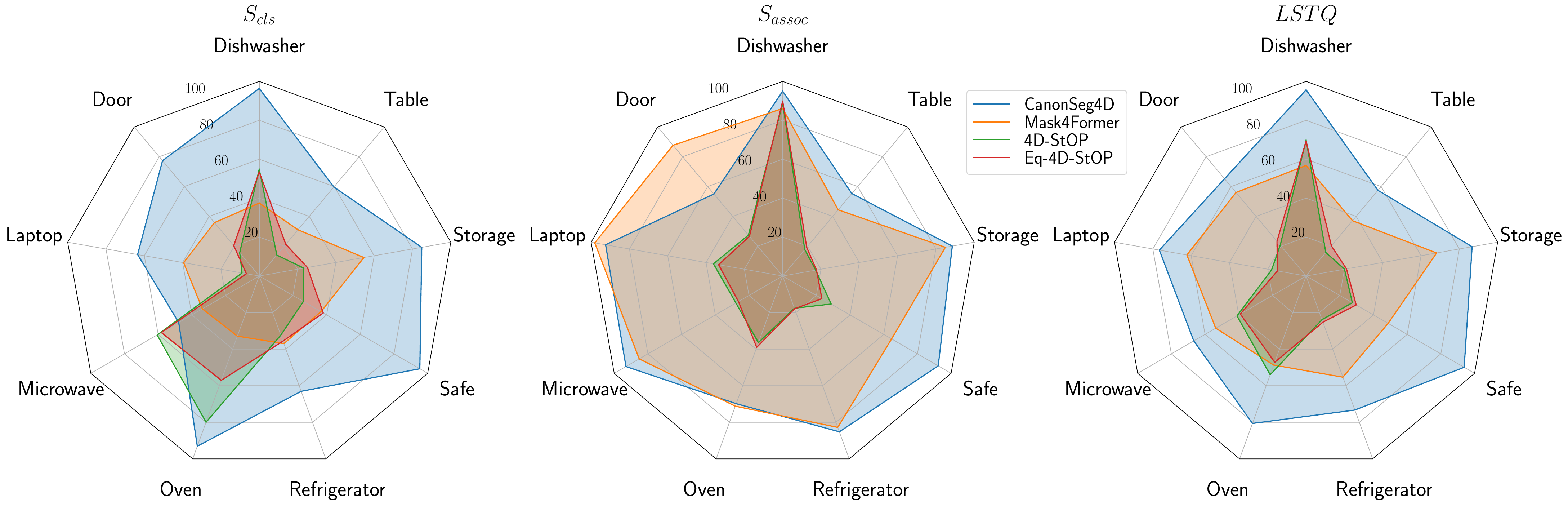}
  \caption{Radar plots comparing performance of the four methods (4D-StOP, Eq-4D-StOP, Mask4former, and CanonSeg4D) across different object categories in the Artic4D-M dataset: semantic segmentation performance ($S_{cls}$), instance association quality ($S_{assoc}$), and combined panoptic segmentation metric ($LSTQ$). Higher values indicate better performance.}
  \label{fig:radar-plot}
\end{figure}

Figure~\ref{fig:radar-plot} provides a category-by-category comparison of all methods on the Artic4D-M dataset across three key metrics.
For semantic segmentation ($S_{cls}$), CanonSeg4D demonstrates clear superiority across most object categories.
The highest performance is achieved on objects with predictable and consistent motion patterns, such as Dishwashers, Ovens, and Safes.
These categories exhibit low intra-class variance, meaning different instances of the same category behave similarly.
Conversely, categories with lower performance typically have either high intra-class variance, like Tables, which encompass various desk types with different configurations of doors and drawers, or limited training samples, such as Microwaves.
The difficulty of these categories is confirmed by the similarly poor performance of competing methods.
Regarding instance association ($S_{assoc}$), while CanonSeg4D remains the overall best performer, Mask4Former shows competitive results in several categories.
Interestingly, categories like Laptops and Microwaves that performed poorly in semantic segmentation show relatively better performance in instance association.
This supports the hypothesis that the semantic confusion occurs primarily between similar functional parts, e.g., doors versus lids, while the physical boundaries between different instances remain distinguishable.
The combined metric ($LSTQ$) demonstrates CanonSeg4D's comprehensive advantage across all object categories, with Mask4Former consistently placing second.
This pattern confirms that CanonSeg4D's balanced performance in both semantic and instance segmentation translates to superior overall 4D panoptic segmentation across diverse articulated objects.

\subsection{Parameter Sensitivity Analysis}

\begin{figure}[t!]
  \centering
  \begin{minipage}[t]{0.48\columnwidth}
    \centering
    \includegraphics[width=\linewidth]{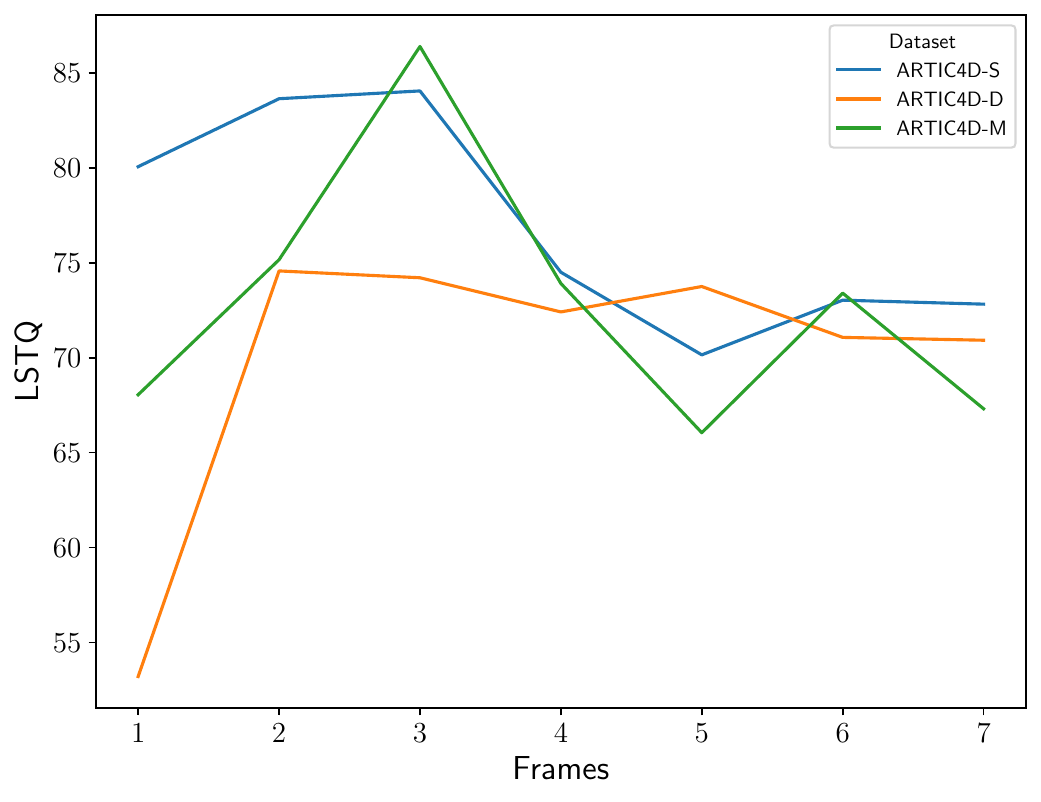}
    \caption{Impact of input sequence length on CanonSeg4D's performance.
      The plot shows the $LSTQ$ score as a function of the number of frames per sequence, evaluated on the three subsets of the Artic4D dataset.
    Performance peaks at a sequence length of three frames across all subsets.}
    \label{fig:frames-line-plot}
  \end{minipage}
  \hfill
  \begin{minipage}[t]{0.48\columnwidth}
    \centering
    \includegraphics[width=\linewidth]{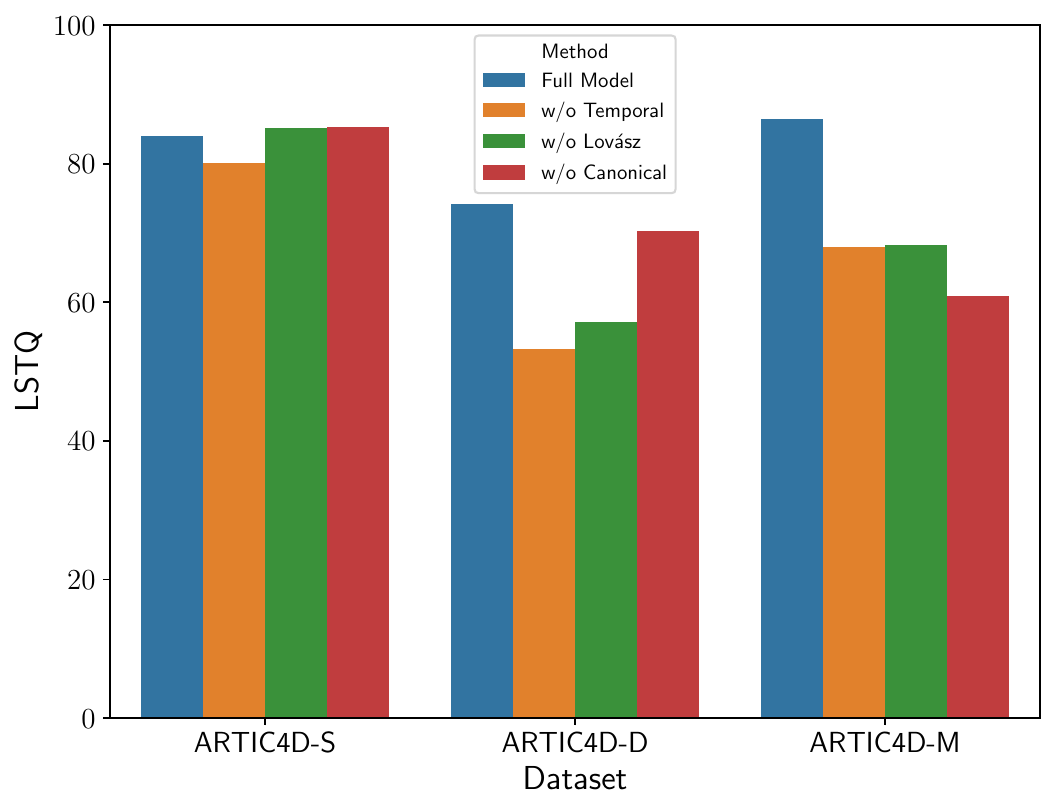}
    \caption{Ablation study results comparing the performance of CanonSeg4D and three variants: without temporal aspect, without Lovász-Softmax loss, and without Canonical Module, showing the impact on $LSTQ$ metrics across the three Artic4D dataset subsets.}
    \label{fig:ablation-results}
  \end{minipage}

  \vspace{0.5em}
\end{figure}
Given the established importance of temporal information, the optimal input sequence length for CanonSeg4D was assessed.
Figure~\ref{fig:frames-line-plot} illustrates the relationship between the number of frames per sequence and the resulting $LSTQ$ score across the three Artic4D subsets.
We can appreciate that a clear trend emerges, as performance with a single frame is severely limited, particularly on the more complex Artic4D-D and Artic4D-M subsets, confirming that static snapshots provide insufficient information to model articulated motion.
A substantial performance increase is observed when the sequence length is increased to three frames, where the model achieves its peak performance across all datasets.
Interestingly, further increasing the sequence length to five and seven frames leads to a noticeable decline in performance, which then plateaus.
This suggests that while temporal context is crucial, longer sequences introduce redundant information or noise that the model struggles to utilize effectively.
Based on this empirical evidence, a sequence length of three frames was identified as the optimal configuration, as it provides a sufficient temporal window to model articulation without introducing the confounding effects of longer, more complex sequences.

\subsection{Ablation Studies}

To isolate the contribution of each architectural component, an ablation study was conducted.
This study systematically evaluates the impact of three key elements: the temporal feature extraction, the Lovász-Softmax loss function, and the Canonical Module.
For the temporal aspect ablation, each frame was processed individually rather than as a sequence.
When removing the Lovász-Softmax loss, it was substituted with the standard Cross-Entropy loss function.
For the Canonical Module ablation, the learned canonical representation was replaced with centroid-based transformation, similar to existing approaches, where points are transformed to their instance centroid.

Figure~\ref{fig:ablation-results} illustrates how removing each component affects performance across the Artic4D dataset subsets.
On the simplest dataset, Artic4D-S, all variants achieved comparable performance, with only a slight decrease when removing temporal features.
This suggests that simpler articulation patterns can be effectively handled even without all components fully engaged.
For Artic4D-D, removing either the temporal aspect or the Lovász-Softmax loss caused significant performance drops.
This highlights that temporal information becomes crucial for understanding more complex movements, while the specialized loss function effectively addresses class imbalance issues highlighted by the higher difficulty of this scenario.
Interestingly, removing the Canonical Module had minimal impact at this complexity level.
The most substantial differences appeared in Artic4D-M, where the complete model significantly outperformed all ablated variants.
The removal of the Canonical Module caused the most dramatic performance decrease, confirming the hypothesis that canonical transformations provide consistent reference configurations that enable better generalization across diverse articulated objects.
The temporal features and specialized loss function also proved essential for handling the complex articulation patterns present in this challenging subset.

\begingroup
\setlength{\tabcolsep}{1.5pt}
\begin{table*}[t!]
  \centering
  \caption{
    Side-by-side comparison of ground truth, CanonSeg4D without the Canonical Module, and the full CanonSeg4D.
    A dishwasher, a cabinet with two doors, and a cabinet with a door and a drawer are shown in the first, second, and third rows, respectively.
    For each method, both the instance segmentation and the transformed point cloud are shown.
    The transformed point cloud refers to the centroid-based representation for the ablated model and the canonical representation for the proposed method, both with the object body removed.
    Different instances are shown in different colors, while different positions of the same instance are shown in different shades of the same color.
  }
  \begin{tabular}{ccccc}
    \toprule
    \multirow{3.5}{*}{\shortstack{Ground\\Truth}} & \multicolumn{2}{c}{W/o Canonical Module} & \multicolumn{2}{c}{CanonSeg4D} \\ \cmidrule{2-5}
    & \multirow{2}{*}{\shortstack{Instance\\Segmentation}} & \multirow{2}{*}{\shortstack{Transformed\\Point Cloud}} & \multirow{2}{*}{\shortstack{Instance\\Segmentation}} & \multirow{2}{*}{\shortstack{Transformed\\Point Cloud}} \\
    & & & & \\
    \midrule
    \includegraphics[width=0.12\textwidth]{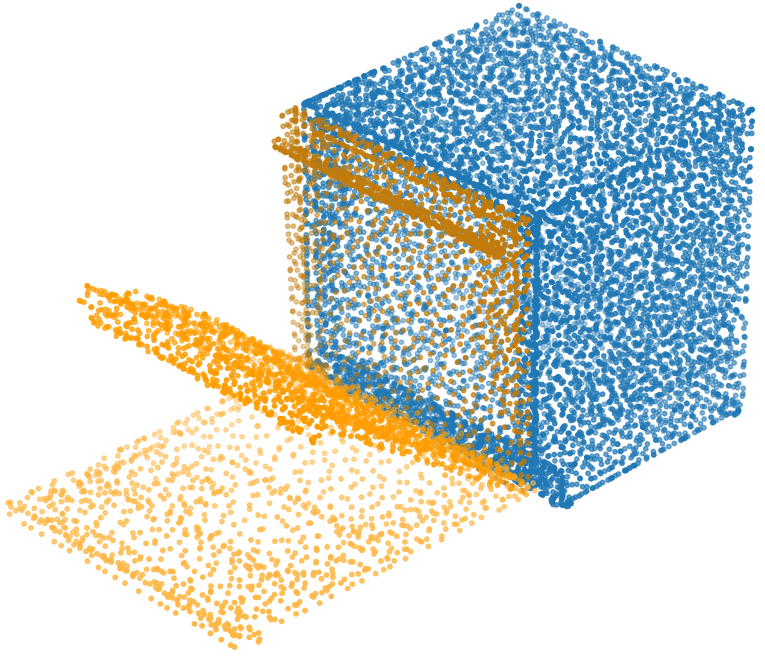} & \includegraphics[width=0.12\textwidth]{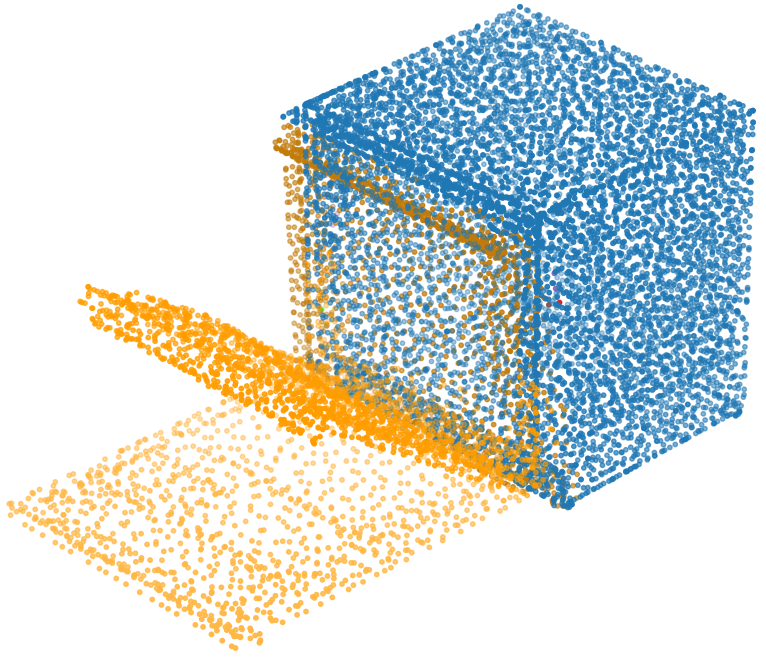} & \includegraphics[width=0.12\textwidth]{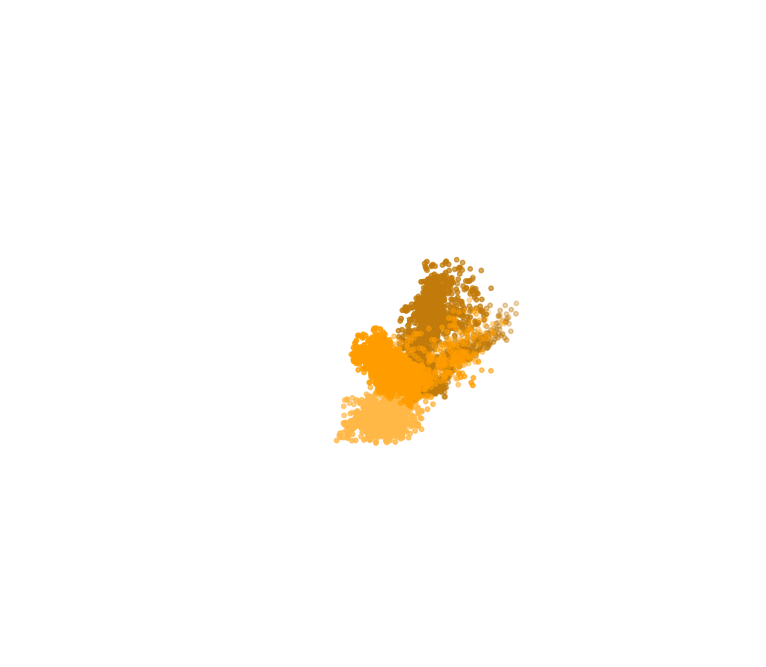} & \includegraphics[width=0.12\textwidth]{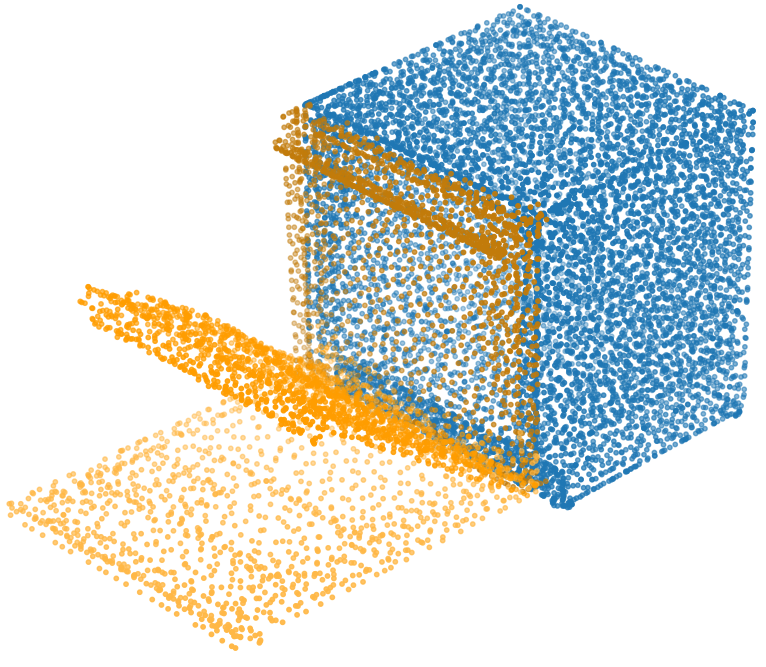} & \includegraphics[width=0.12\textwidth]{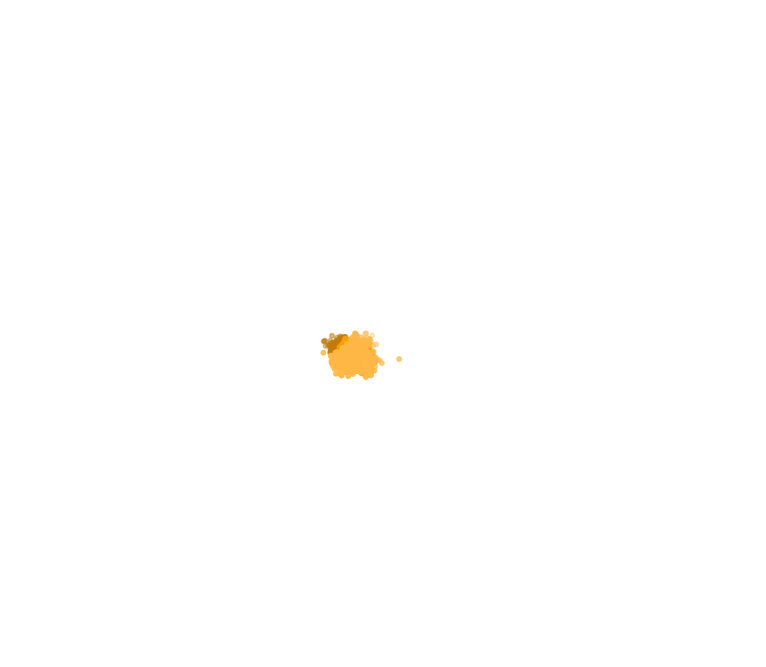} \\
    \includegraphics[width=0.12\textwidth]{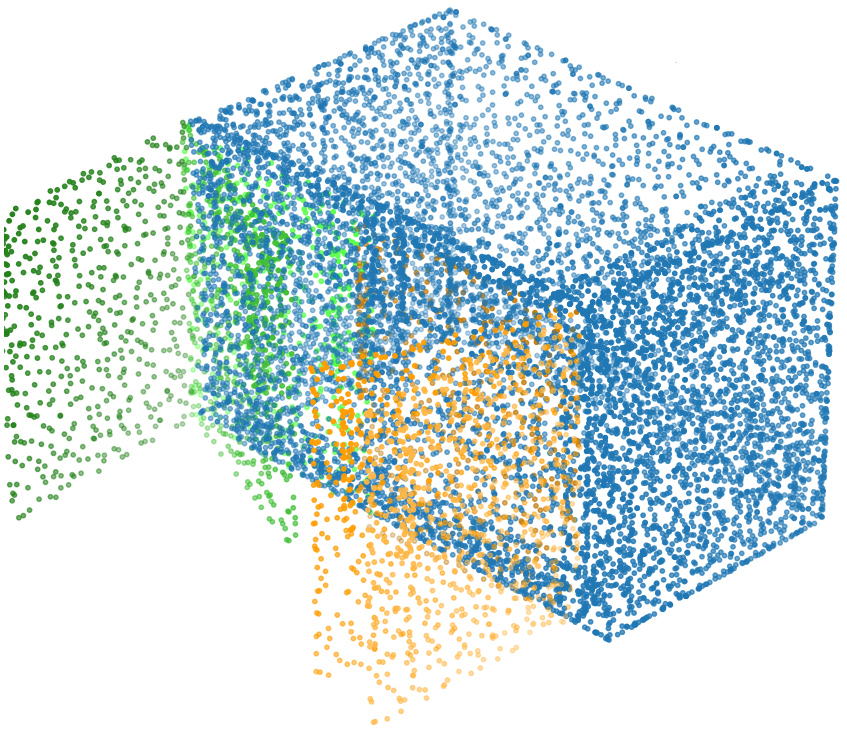} & \includegraphics[width=0.12\textwidth]{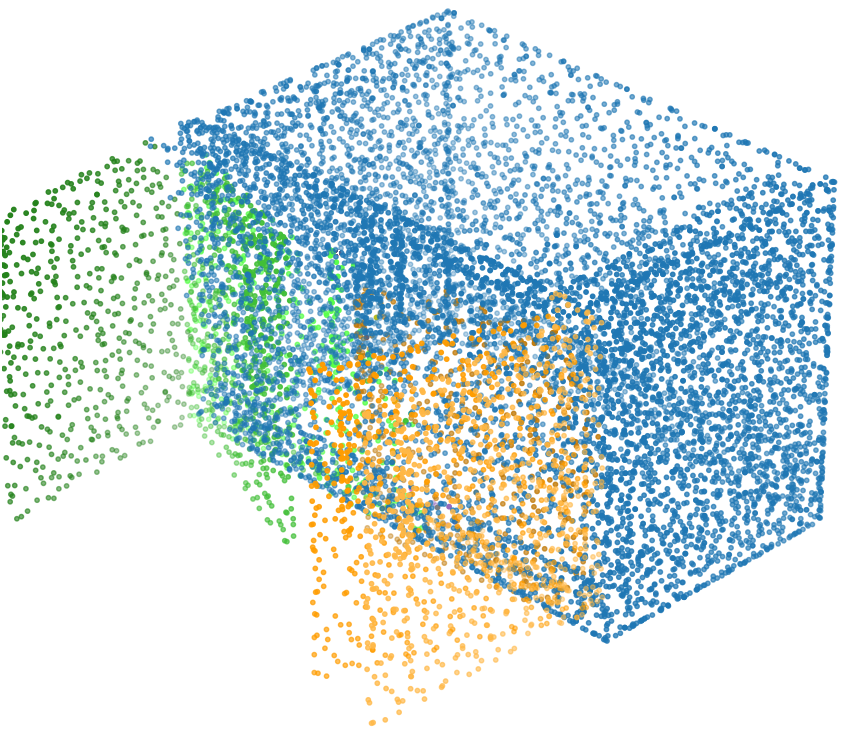} & \includegraphics[width=0.12\textwidth]{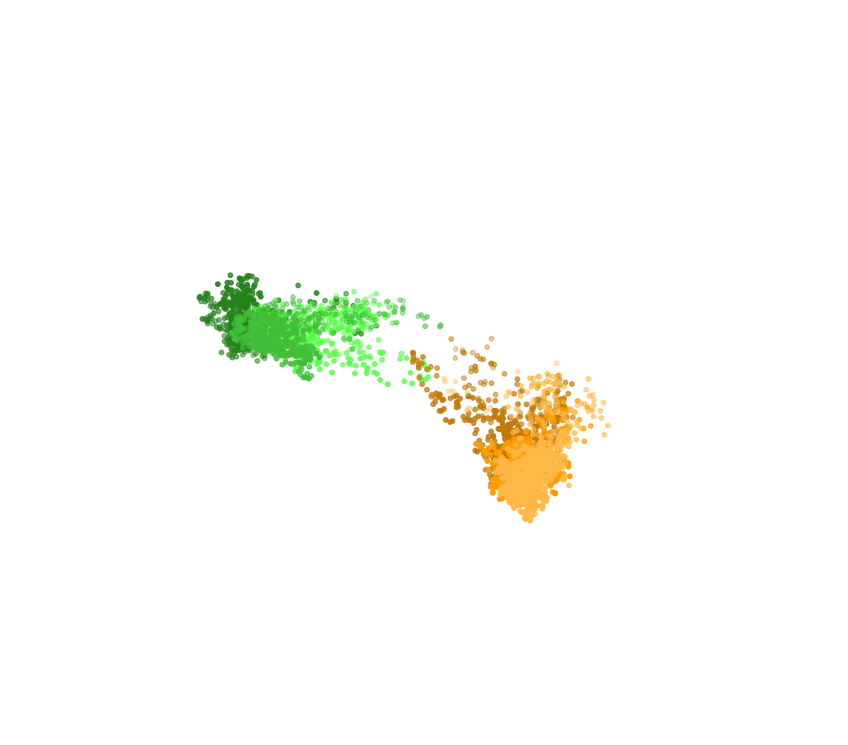} & \includegraphics[width=0.12\textwidth]{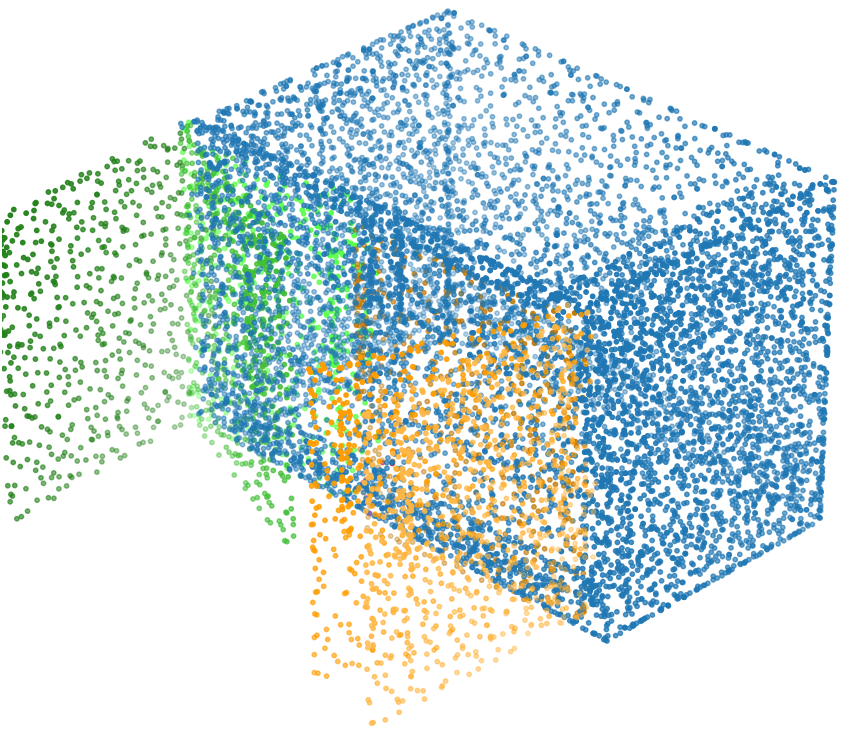} & \includegraphics[width=0.12\textwidth]{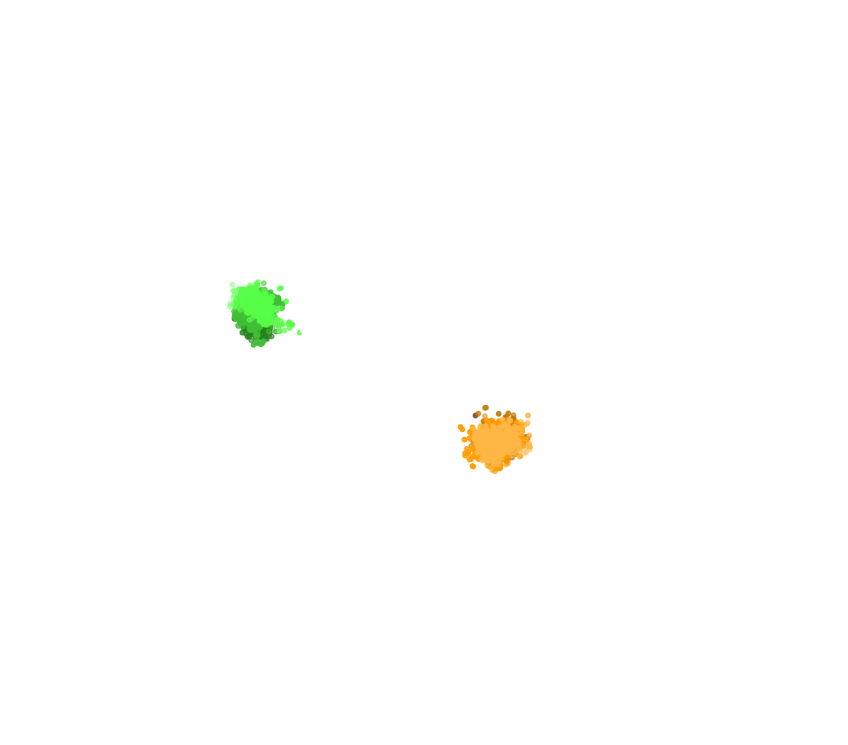} \\
    \includegraphics[width=0.12\textwidth]{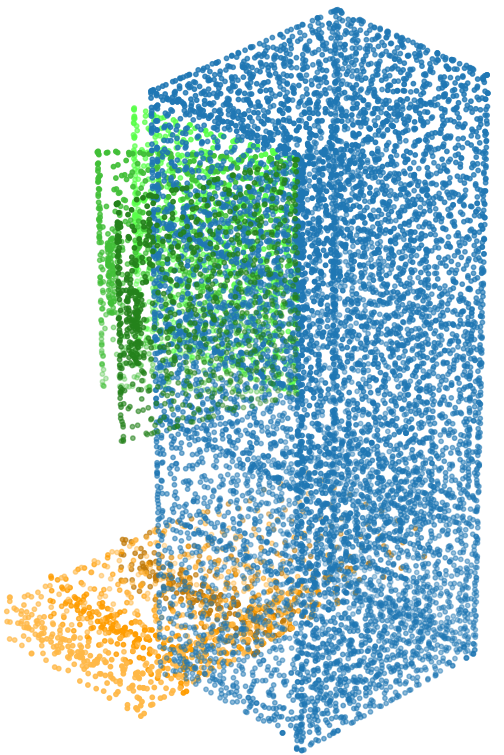} & \includegraphics[width=0.12\textwidth]{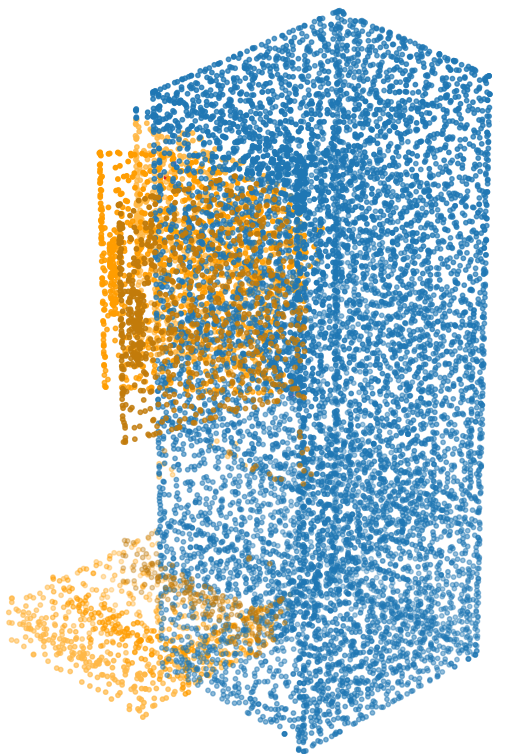} & \includegraphics[width=0.12\textwidth]{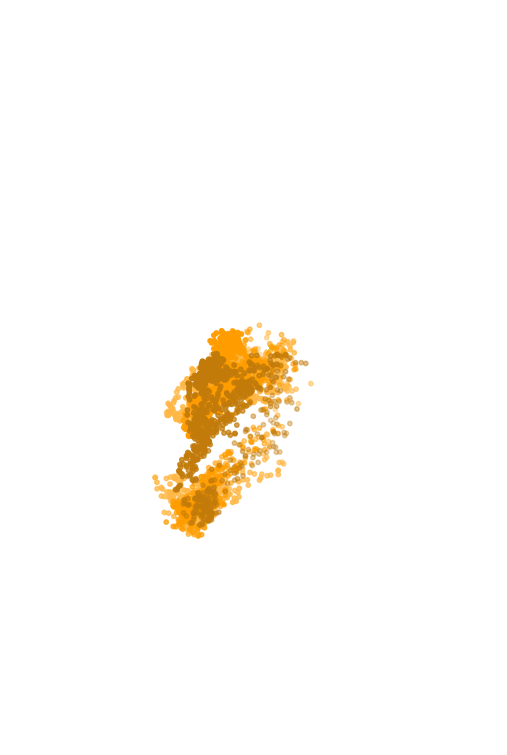} & \includegraphics[width=0.12\textwidth]{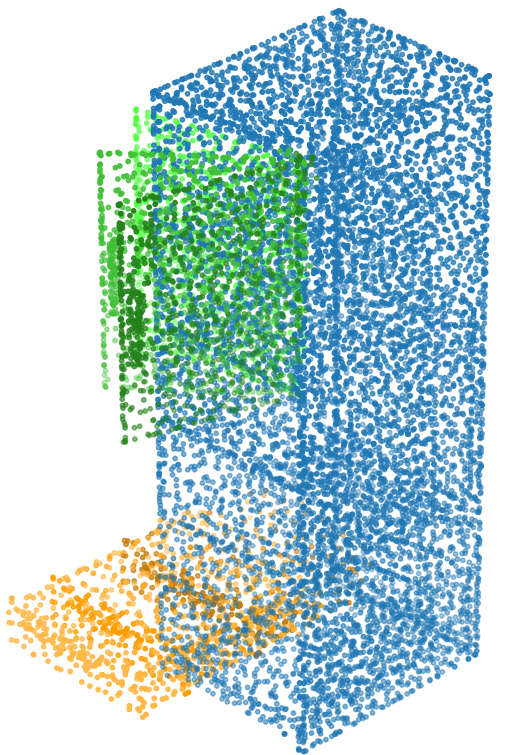} & \includegraphics[width=0.12\textwidth]{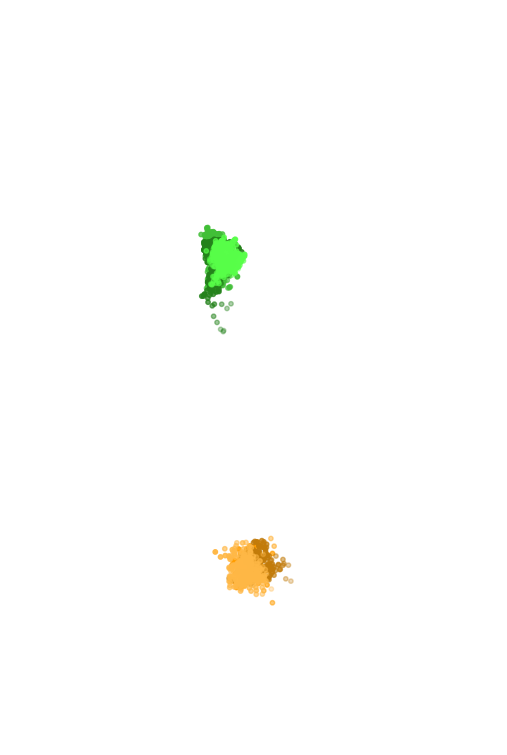} \\
    \bottomrule
  \end{tabular}
  \label{tab:qualitative-comparison}
\end{table*}
\endgroup

Table~\ref{tab:qualitative-comparison} further highlight the importance of the Canonical Module, the main novelty of this approach.
It provides a visualization of the results on the Artic4D-M dataset, illustrating the performance of CanonSeg4D against the ablated version without the Canonical Module.
The comparison focuses solely on instance segmentation, as this is where the difference between the baseline and ablated model lies.
In the first example, a simple object with one moving part, both methods correctly segment the instance.
However, the proposed method's transformed point cloud is visibly more consistent.
This stability stems from the learned canonical space, which provides a regularized target that improves the reliability of the transformation learning process.
The second example, featuring an object with two moving doors, further demonstrates this advantage.
While both methods segment the doors correctly, the clusters produced by the ablated model are irregular and nearly overlap.
This proximity introduces ambiguity, making it difficult to distinguish between the two parts.
In contrast, the proposed method generates well-defined, clearly separated clusters.
In the third example, the complete failure of instance segmentation is directly attributable to cluster merging within the ablated model, which prevents the two moving parts from being differentiate.
The proposed method, however, maintains distinct and well-separated clusters for each part.

\section{Discussion}
Artic4D, used as a benchmark dataset, enables clear, reliable, and reproducible evaluation, shown by making  both ablation studies and state-of-the-art comparisons straightforward, supporting apples-to-apples experiments with minimal effort.
The dataset is publicly available upon acceptance, facilitating community benchmarking and fair comparisons.
The three subsets are well designed, as they delineate increasing levels of difficulty.
This stratification supports more fine-grained analysis and yields deeper insights into how methods behave as semantic diversity and motion complexity grow.

A key limitation is coverage: given the breadth of PartNet Mobility, we instantiated Artic4D with only the most commonly used categories, leaving various object models untested.
Although the same generation pipeline is already used across categories, suggesting similar quality for the unseen ones, this assumption lacks empirical validation.
This uncertainty may impact scalability for more extensive versions of the dataset.

Our approach, CanonSeg4D, takes full advantage of the canonical space transformations, which showed tighter, more discriminative clusters as object variability increases.
The Lovász-Softmax loss proves effective for articulated objects, mitigating the class imbalance typical of this type of object.
Incorporating temporal features further enhances segmentation, suggesting that current state-of-the-art methods for articulated object perception underuse the intrinsic dynamics of these objects.
Compared with 4D panoptic segmentation state-of-the-art approaches, competing methods were less adaptable to this setting, and CanonSeg4D achieved a clear lead in performance.

One limitation is specialization: the method targets articulated objects and assumes a meaningful canonical space; it does not transfer easily to motions that defy parametric descriptions.
Suitable scenarios outside articulated objects where canonicalization applies are scarce, making broader benchmarking difficult.
Another limitation is that canonical supervision adds annotation overhead: while reliable in simulation, obtaining consistent canonical references in real captures is challenging due to noise, occlusion, and alignment ambiguity, increasing data collection cost and complexity.
\section{Conclusion}
This work introduced Artic4D and its modular processing pipeline to standardize articulated object perception research.
The dataset enables direct, fair comparisons across approaches and is designed for straightforward extension.
We also present CanonSeg4D, the first 4D panoptic segmentation method explicitly tailored to articulated objects.
The approach leverages canonical space representations, temporal feature aggregation, and a task-specific semantic loss to deliver consistent, articulation-invariant segmentation over time.

Extensive experiments on Artic4D show that CanonSeg4D outperforms strong state-of-the-art approaches, particularly in the most challenging settings, providing robust semantic and instance segmentation with improved temporal consistency.
Ablation studies confirm the complementary benefits of canonical-space transformations, the specialized semantic loss, and the exploitation of temporal features.
A parameter analysis also identifies an effective temporal length for the 4D point cloud.

Looking ahead, the Artic4D dataset and its modular pipeline provide a strong foundation for future research, enabling the community to extend, adapt, and build variants of the dataset according to their needs.
However, expansion should be done with caution; adding new categories should include a validation protocol to verify fidelity before inclusion in new dataset versions.
One promising avenue is to integrate 4D panoptic segmentation with higher-level perception tasks, such as joint localization, articulation parameter estimation, and manipulation planning, which could enable a full pipeline from sensor data to robotic manipulation, yielding holistic articulated object perception.
Another avenue is to relax the dependence on manually defined canonical spaces.
Future work could learn canonicalization directly from data with weak or self-supervision, jointly inferring kinematic structure and canonical states, which would ease transfer to motions that do not follow parametric rules.
Evaluations on more complex objects and interactions (closed-chain mechanisms and multi-object scenes) will help characterize failure modes and drive more general solutions.

\section*{Acknowledgments}
This work was supported by FCT - Fundação para a Ciência e Tecnologia, I.P. under unit 00127-IEETA by 2023.01882.BD and \href{https://doi.org/10.54499/2023.01882.BD}{https://doi.org/10.54499/2023.01882.BD} and by Grant PID2022-143257NB-I00 funded by MCIN/AEI/10.13039/501100011033
and by "ERDF A Way of Making Europa", the Departament de Recerca i Universitats from Generalitat de Catalunya with reference2021SGR01499, and the Generalitat de Catalunya CERCA Program.
\bibliographystyle{apacite}
\bibliography{references}

\end{document}